\UseRawInputEncoding
\documentclass[final,3p,times,twocolumn]{elsarticle}

\usepackage{xcolor}
\usepackage{amssymb}
\usepackage{natbib}
\usepackage[normalem]{ulem}
\useunder{\uline}{\ul}{}
\usepackage{tabularx}
\usepackage{booktabs}
\usepackage{adjustbox}
\usepackage{booktabs}
\usepackage{siunitx}
\usepackage{graphicx}
\usepackage{subcaption}
\usepackage{algorithm}
\usepackage{algpseudocode}
\usepackage{changepage}
\usepackage{url}
\usepackage{breakurl}  
\usepackage{hyperref} 
\hypersetup{breaklinks=true}
\usepackage{amssymb}  
\usepackage{textcomp}
\usepackage{amsmath}

\journal{Energy Reports}

\begin{document}

\begin{frontmatter}



\title{Optimizing Electric Vehicle Charging Station Placement Using Reinforcement Learning and Agent-Based Simulations}

\author[cei,cecs]{Minh-Duc Nguyen}
\ead{duc.nm2@vinuni.edu.vn}

\author[cecs]{Dung D. Le}
\ead{dung.ld@vinuni.edu.vn}

\author[cei]{Phi Long Nguyen\corref{cor1}}
\ead{long.np2@vinuni.edu.vn}
\cortext[cor1]{Corresponding author}

\affiliation[cei]{organization={Center for Environmental Intelligence, VinUniversity},
            city={Hanoi},
            country={Vietnam}}

\affiliation[cecs]{organization={College of Engineering $\&$ Computer Science, VinUniversity},
            city={Hanoi},
            country={Vietnam}}

\begin{abstract}
The rapid growth of electric vehicles (EVs) necessitates the strategic placement of charging stations to optimize resource utilization and minimize user inconvenience. Reinforcement learning (RL) offers an innovative approach to identifying optimal charging station locations; however, existing methods face challenges due to their deterministic reward systems, which limit efficiency. Because real-world conditions are dynamic and uncertain, a deterministic reward structure cannot fully capture the complexities of charging station placement. As a result, evaluation becomes costly and time-consuming, and less reflective of real-world scenarios. To address this challenge, we propose a novel framework that integrates deep RL with agent-based simulations to model EV movement and estimate charging demand in real time. Our approach employs a hybrid RL agent with dual Q-networks to select optimal locations and configure charging ports, guided by a hybrid reward function that combines deterministic factors with simulation-derived feedback. Case studies in Hanoi, Vietnam, show that our method reduces average waiting times by 53.28$\%$ compared to the initial state, outperforming static baseline methods. This scalable and adaptive solution enhances EV infrastructure planning, effectively addressing real-world complexities and improving user experience.
\end{abstract}


\begin{graphicalabstract}
\includegraphics[scale = 0.5]{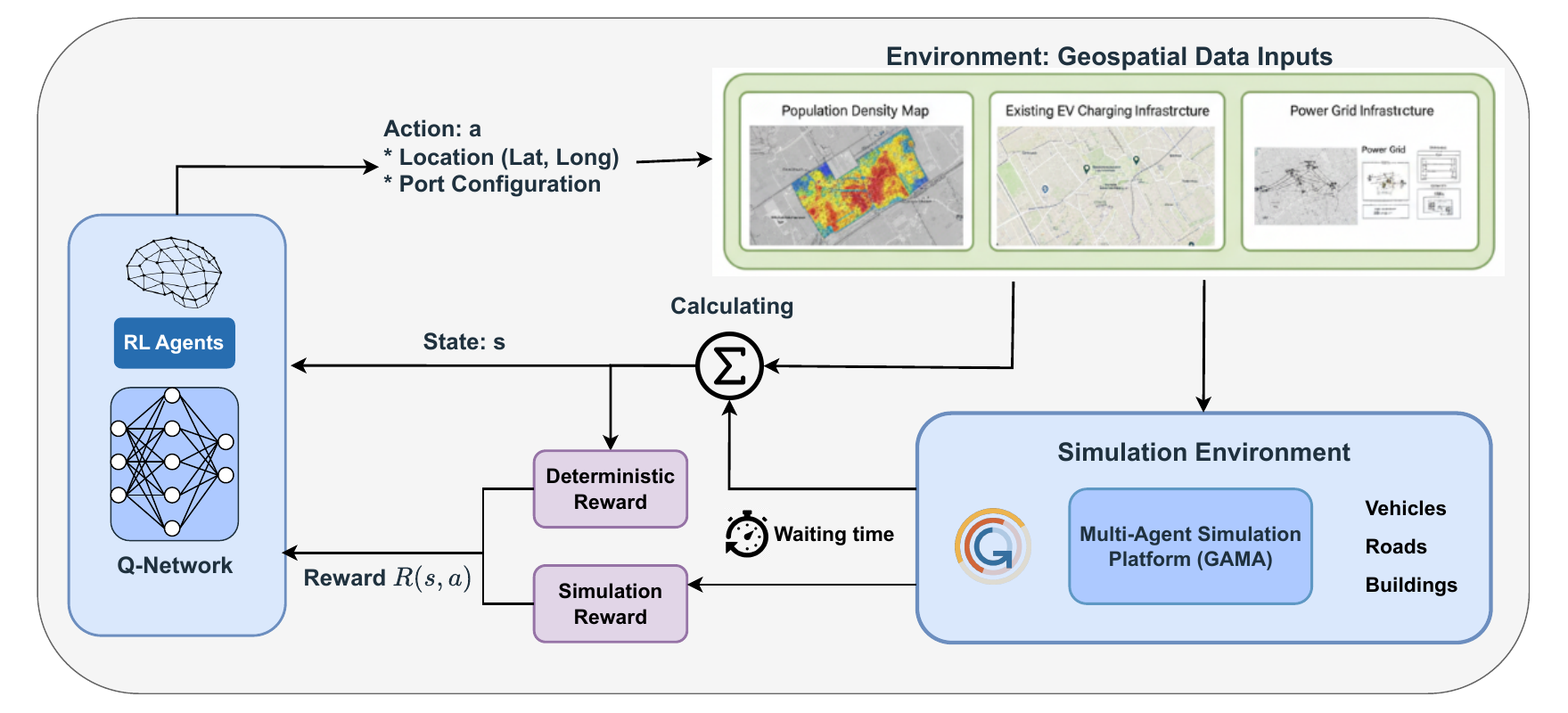}
\end{graphicalabstract}

\begin{highlights}

\item Introduce an innovative framework that incorporates reinforcement learning (RL) to choose the optimal location for the charging station. We present the novel action space in the RL algorithm, adapting to the charging station system.

\item  Integrate a simulation system to replicate the movement of the Electric Vehicle (EV) in the city, aiming to obtain feedback from the virtual environment immediately and decide on the next location for charging station placement.

\item We design a simulation reward that is flexible, reflecting the responses from the environment. This reward allows the model to be more suitable in reality instead of only depending on the deterministic reward.

\item Our method achieves a 53.28\% reduction in average waiting times for electric vehicle charging, outperforming static baseline methods. This study demonstrates the effectiveness of the RL-based method along with the simulation system.    


\end{highlights}


\end{frontmatter}


\section{Introduction}
\label{sec:introduction}
Climate change is a critical global issue with profound environmental impacts on human civilization and long-term survival. Its primary drivers include harmful human activities such as untreated domestic and industrial wastewater and excessive CO2 emissions from fossil fuel combustion. Proposed solutions emphasize adopting clean energy sources to reduce carbon emissions. In line with this, major cities worldwide are transitioning toward green cities by promoting electric vehicles (EVs). To support this transition, a robust network of charging stations is essential, tailored to the specific needs and characteristics of each region. Establishing such a network requires addressing complex factors that influence optimal placement.

Advances in modern technology have enabled the collection of vast amounts of environmental data from multiple sources, providing valuable insights for selecting optimal charging locations. While comprehensive environmental information offers clear benefits, managing and processing large datasets present considerable challenges. Identifying suitable locations for electric vehicle charging stations (EVCS) therefore, requires more sophisticated methods than traditional linear programming approaches. Multiple factors, including land use patterns, population density, and traffic flows, must be integrated into the analysis, making it a complex data-mining task.

Addressing the EVCS site selection challenge necessitates the integration of data from a wide range of sources. These may be spatial or non-spatial, qualitative or quantitative, and may range from discrete values to nearly immeasurable figures. Recent research has made significant strides in addressing this complexity through geospatial analysis using Geographic Information Systems (GIS) software. GIS utilizes spatial data to analyze and determine the optimal locations for charging stations \cite{10.1007/978-981-19-2273-2_8,en17184546}. A comprehensive review of 74 studies from 2010 to 2023 highlights the effectiveness of GIS-based methods for EVCS location optimization \cite{Banegas2023}. These studies demonstrate that GIS techniques, including map algebra and data overlay methods, are particularly effective at integrating multiple selection criteria into the decision-making process.

The placement of EV charging stations (EVCS) is often modeled as a multiple criteria decision-making (MCDM) problem, considering factors like proximity to major roads, population density, land cost, traffic volume, and power availability. To support the balancing of these factors, Geographic Information Systems (GIS) provide spatial mapping and weighting of these factors to aid balanced decision-making. Furthermore, methods such as the Analytical Hierarchy Process (AHP) \cite{su13042298, MHANA2024105456, HUANG2019113855} and integrated GIS-MCDM approaches, including Multi Influencing Factor (MIF) weighting and TOPSIS ranking \cite{RANE2023104717}, have been widely adopted \cite{IRAVANI2022100135, su151410967,en14102756, ERBAS20181017,KAYA2020102271}.

However, these approaches heavily rely on expert-defined parameters and static datasets, making them vulnerable to inaccuracies and limiting their scalability. The absence of real-time data, such as dynamic traffic patterns and energy demand, further constrains their adaptability and reduces their effectiveness in rapidly changing urban environments.

Machine learning techniques are increasingly used to predict optimal locations for electric vehicle charging stations by forecasting demand, optimizing site selection, and addressing spatial disparities \cite{SREEKUMAR2024103095,sanami2025demandforecastingelectricvehicle,BELLIZIO2025101657,su15032105}. In the current study, deep reinforcement learning (DRL) offers adaptive decision-making by accounting for dynamic factors like traffic flow, station usage, and grid constraints. For example, an advantage actor-critic (A2C) DRL model \cite{HEO2024105567} was proposed to optimize EV fast charging station placement using integrated geospatial data. However, these models often rely on static environments and manually designed reward functions; specifically, the reward is calculated deterministically from static factors, limiting their ability to fully capture real-world dynamics and variability.

In this study, we address these limitations by employing a simulation system that dynamically evolves over time, capturing the flexibility and variability of real-world environments. 
Current evaluation methods predominantly rely on retrospective analysis, utilizing historical data to evaluate factors such as charging efficiency, station utilization, and overall system reliability. However, these approaches do not adequately capture dynamic fluctuations in demand, charging delays, and user experience in real time. Recent work \cite{JORDAN2022116739} employs an agent-based simulation to estimate the waiting time metric, capturing how congestion and station availability influence user experience. This approach is particularly valuable as it allows for scenario testing, optimization of resource allocation, and evaluation of system resilience under varying conditions. By leveraging a simulation system, we can gain valuable insights into user demand patterns, charging behavior, and system utilization. Additionally, by incorporating real-world data into these simulations \cite{LEE202053}, we can refine demand forecasting models, optimize station placement, and improve overall system efficiency.

Building on the points discussed above, we propose an innovative framework that integrates a reinforcement learning (RL) algorithm with a dynamic simulation system to recommend optimal charging station placements. In this framework, an agent is tasked with identifying new charging station locations, while the simulation environment continuously updates and provides real-time feedback to guide the agent's decisions. Specifically, our main contributions are as follows:

\begin{itemize}
   \item We propose a novel framework for determining the optimal locations of electric vehicle charging stations by leveraging deep reinforcement learning (DRL) and integrating the charging demand simulation using the Gama model in the learning process. Our approach captures the dynamic nature of user demand and traffic patterns from simulation, enabling adaptive and efficient station placement.
    \item We present a comprehensive framework for simulating the movement of electric vehicles using GAMA software. This framework enables the modelling of real-world traffic dynamics, charging station interactions, and user behaviors in a controlled virtual environment. By leveraging GAMA's agent-based modelling capabilities, we can simulate individual EV movements and evaluate the impact of charging station availability.
    \item We conducted extensive experiments in Hanoi, Vietnam, to validate the performance and robustness of our proposed framework under real-world conditions. Through these experiments, we systematically analyzed key performance metrics, including waiting times and overall system efficiency. Our results demonstrate significant improvements, particularly in reduced waiting times compared to baseline approaches.
\end{itemize}

This paper is organized as follows: Section 2 reviews related work, Section 3 presents our framework, Sections 4 and 5 detail experiments and results, and Section 6 concludes.

\section{Related Work}
\subsection{Traditional Optimization Techniques.}
Traditional Optimization Techniques
Traditional optimization techniques have played a crucial role in addressing the problem of Electric Vehicle Charging Station (EVCS) placement. These methods typically involve formulating the problem as a mathematical model and solving it using approaches like linear programming, integer programming, or greedy algorithms. Research suggests that these approaches are effective for static scenarios but may struggle with dynamic demand. Key developments in this area include comprehensive summaries of optimization techniques, including linear programming and set-covering models, emphasizing their importance in determining the optimal placement and size of EVCS \cite{Islam20215}; studies highlighting the importance of integrating power grid constraints into placement strategies, demonstrating how traditional methods address infrastructure limitations \cite{AHMAD20222314}. The distribution network (DN) provides the electric power, which significantly impacts the new charging location \cite{zeb2020optimal, pal2021allocation}. The formulations of the EVCS placement problem as a complex optimization challenge propose solutions like greedy algorithms to balance computational feasibility and solution quality \cite{lam6688009}. In general, the EVCS problem formulates the objective function based on many factors from the environment, and many approaches depend on user behaviour or the goal of multiple stockholders. The cost has been considered an objective function like investment cost, operating and
maintenance cost \cite{kazemi2016optimal}; traveling cost, operation cost \cite{xiang2016economic,zhang2015integrated}; construction cost \cite{luo2020electric,ren2019location}. These studies demonstrate that traditional optimization techniques are robust for initial planning and well-suited to static scenarios. However, they often lack adaptability to dynamic conditions, such as fluctuating EV demand or real-time grid constraints.

\subsection{Heuristic and Metaheuristic Methods}
Heuristic and metaheuristic methods address the complexity and non-linearity of the EVCS placement problem by offering flexible and scalable solutions, particularly for large-scale or real world applications. These approaches trade strict optimality for computational efficiency. Key contributions include applications of genetic algorithms (GA) to minimize costs based on detailed cost models, optimizing placement in resource-constrained environments \cite{ZHOU2022123437}; research employing GAs to determine optimal EVCS locations in urban settings, emphasizing coverage and accessibility \cite{Efthymiou2017}; hybrid approaches combining GAs with simulated annealing to enhance placement resilience against distribution network disruptions \cite{Kumar2024}; and studies using GAs to simultaneously minimize costs and maximize coverage, illustrating the versatility of metaheuristic approaches \cite{10587625}. Heuristic and metaheuristic methods, particularly genetic algorithms, excel in handling complex, multi-objective problems. However, they do not guarantee globally optimal solutions and often require careful parameter tuning to achieve satisfactory results.
\subsection{Machine Learning and Data-Driven Approaches} Machine learning (ML) and data-driven approaches leverage large datasets, such as traffic patterns, EV usage, and demographic data, to predict demand and inform EVCS placement. These methods enhance decision-making by identifying patterns that traditional techniques might overlook. Notable approaches include clustering techniques to group areas of high EV demand, guiding placement decisions \cite{systems10010006}; integrations of ML to predict demand and optimize both placement and scheduling in urban environments \cite{su152216030}; ML models forecasting the spatial distribution of EVCS, aiding long-term planning \cite{Rai10002556}, and analyses addressing inequities in EVCS placement across regions using ML \cite{roy2022examining}; ML-based approaches offer significant predictive power and adaptability to diverse datasets. However, they require extensive data preprocessing and may not directly optimize for specific objectives like cost or grid stability, often serving as a complementary tool rather than a standalone solution. Additionally, probabilistic load forecasting, a critical foundation for EVCS planning, employs advanced ML techniques, such as reinforcement learning (RL)-assisted deep learning, to predict charging power demand with high accuracy, accounting for uncertainties in EV usage \cite{9760076}. RL-based methods further enhance forecasting by modeling complex scenarios, such as hierarchical probabilistic demand prediction \cite{10365547} and dynamic load balancing \cite{mosalli2025dynamic}, complementing siting decisions with real-time operational optimization.

While traditional optimization excels in static scenarios, it lacks adaptability to real-time changes. Heuristic methods offer flexibility but not optimality, and ML approaches depend heavily on data quality. Our RL framework, integrated with GAMA simulations, overcomes these limitations by dynamically adapting to EV demand and traffic patterns, using simulation-based rewards to enhance decision-making over deterministic models.
\section{Deep Reinforcement Learning with Agent-based Simulation}
\subsection{Charging Station Location Problem}
The rapid adoption of electric vehicles (EVs) necessitates the development of an efficient charging infrastructure to accommodate growing demand. A key challenge in optimizing this infrastructure is minimizing the system-wide average waiting time for EV users at charging stations. Our problem concentrates on finding the next station such that the waiting time on our charging system is minimized.
Let $S = \{s_1, s_2, ..., s_n\}$ represent the set of existing charging stations, where n is the total number of existing stations. Each station $s_i$ is located at position $P_i = (x_i, y_i)$. The waiting time associated with each station $s_i$ is denoted as  $w_{s_i}$, which reflects congestion levels and service availability at that location.

The primary objective of this study is to determine the optimal placement of a new charging station $s_{n+1}$ that minimizes the overall system-wide waiting time for EV users. Mathematically, this objective can be formulated as follows:

\begin{equation}
    \min_{P_{n+1}} W_{total} = \sum_i^{n}w_{s_i} + w_{s_{n+1}}
\end{equation}
where $w_{s_{n+1}}$ represents the waiting time at the newly introduced charging station $s_{n+1}$. 
\begin{figure*}[!t]
    \centering
    \includegraphics[width=1\linewidth]{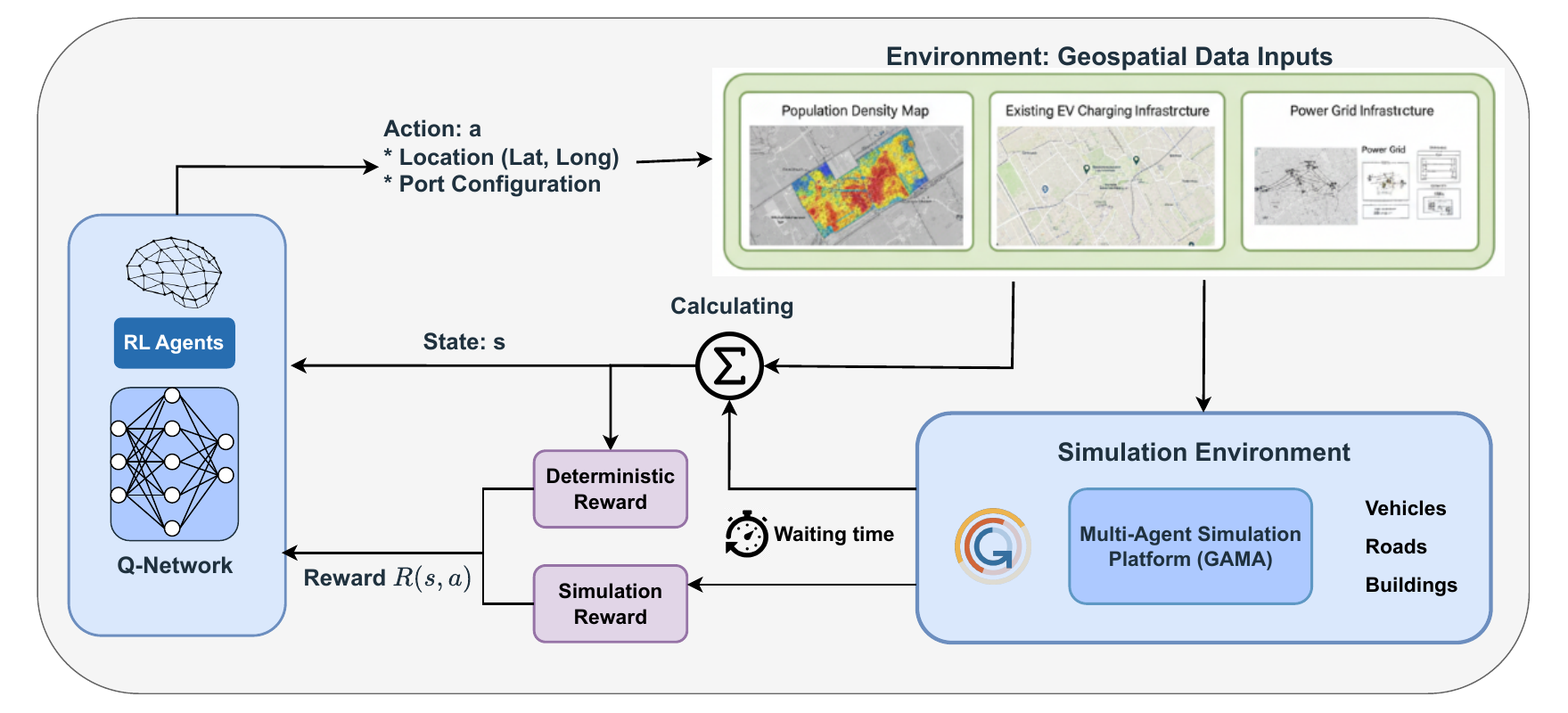}
    \caption{Workflow of Charging Station Placement Agent with Environment and Simulation Feedback}
    \label{fig: framework}
\end{figure*}
\subsection{Reinforcement Learning Framework for Charging Station Deployment}
We determine the optimal location for the next charging station with the objective of minimizing overall waiting time in the future. At each step, we thoroughly assess potential locations by considering key factors such as population density and the existing distribution of charging stations and electric stations. This approach closely mirrors the reinforcement learning (RL) paradigm, where the model continuously refines its strategy based on feedback from the environment.

In our methodology, we integrate dynamic variables derived from the simulation environment to enhance decision-making at each stage. Rather than focusing solely on short-term benefits, we also account for the long-term effects on the charging network. This entails a detailed analysis of how a newly established station impacts future traffic patterns, overall system efficiency, and accessibility for users.

To achieve this, we implement an RL-based framework for selecting the next charging station location and utilize the GAMA simulation platform to model the behavior of the charging infrastructure, ensuring that our approach captures real-world dynamics and optimizes network performance.

Our proposed framework consists of three main components shown in Figure \ref{fig: framework}:
\begin{itemize}
    \item \textbf{Charging Station Placement Agent}: The charging station placement agent is responsible for determining optimal locations for new charging stations by learning from environmental feedback and optimizing placement strategies based on received rewards. It processes state information, including population density, proximity to the neighbour charging station, proximity to the nearest electric station, and the waiting time for each station in the nearest station, to make informed decisions that enhance the efficiency and accessibility of the charging network.
    \item \textbf{Environment}: The environment represents the real-world context in which the agent operates. We construct a data-driven environment that reflects both population distribution and the current state of charging station infrastructure. When the agent makes a decision, the environment immediately returns a state that includes information about the surrounding landscape of the selected location. Additionally, the environment computes a reward based on the agent's action. The reward function is carefully designed to capture the relative impact of various environmental factors.
    \item \textbf{Simulation Environment}: The GAMA simulation platform serves as a powerful tool for supporting this system. It offers advanced agent-based simulation capabilities, realistic traffic flow modelling, and user behaviour simulation. We employ a simulation system aimed to reflect the vary of situations of the environment when a new charging station is located. Furthermore, it facilitates accurate waiting time calculations and provides dynamic updates to the system state, enabling adaptive and context-aware decision-making.
\end{itemize}

Integrating the reinforcement learning approach with the simulation system introduces an innovative framework. Typically, the charging station system is simulated only after a new station is selected to assess its effectiveness. However, in our study, we directly incorporate a simulation system that plays a crucial role in shaping the environment for agent movement. Similar to real-world RL, agents in our framework can make decisions based on their flexible environment to maximize rewards.

\subsection{Deep Reinforcement Learning Model}
We model the problem of charging station placement as a reinforcement learning (RL) task. Specifically, when a new charging station is selected, the surrounding environment may change, affecting factors such as traffic density or user charging behaviour. In RL terms, an agent takes an action (selecting a charging station location with its port type configured) and observes the resulting changes in the environment's state. The agent uses this feedback to update its policy for making better future decisions.

In our framework, we refer to this agent as the charging station placement agent. As illustrated in Figure \ref{fig: framework}, the agent selects a point on the map as the location for a new charging station. The environment responds with a new state, including information such as population density and the distances to nearby stations, distances to nearby electric stations, and the average waiting time in this area. A reward is then calculated, combining deterministic and simulation-based components. By learning from this feedback, the agent optimizes its strategy for placing subsequent charging stations, aiming to improve overall network performance.

\subsubsection{State of the Environment}
We carefully account for various factors that influence the selection of a new charging station to define the current state. Previous studies \cite{en14102756,RANE2023104717} identified several key criteria for optimal EVCS placement, including station properties, energy availability, urban characteristics, physiography, financial considerations, and transportation infrastructure. While using multiple criteria provides a comprehensive approach, it also increases model complexity by requiring numerous parameters to quantify the influence of each factor, a notable limitation of these models.

In our framework, we leverage deep learning to incorporate these factors as input data effectively. Specifically, we focus on key variables such as population density, distance to existing stations, proximity to the nearest substation, and availability of sustainable energy to construct the state representation for our model.

The state $s$ represents the environment’s characteristics at a candidate location for a charging station. It is a multidimensional vector that captures spatial, demographic, and operational factors influencing placement decisions. The state vector includes the following six components:

\begin{itemize}
    \item Distance to Nearest Station \( d_{\text{nearest}} \) (km):  
  The geodesic distance from the candidate location \( p \) to the nearest existing station:
  \[
  d_{\text{nearest}} = \min_{i} \text{geodesic}(p, s_i)
  \]  where $s_i$ is the location of the $i$-th existing station. This feature encourages the agent to prioritize underserved areas by favoring locations far from existing infrastructure.
  \item Distance to Nearest Substation \( d_{\text{nearest\_sub}} \) (km): We introduce a variable that allows the agent to observe all surrounding substations. This variable provides essential information about the distribution of substations, enabling the agent to make more informed decisions. In our study, we emphasize that a newly established charging station should be strategically positioned to ensure compatibility with the substation system, facilitating efficient electricity supply and distribution.
   \[
  d_{\text{nearest\_sub}} = \min_{i} \text{geodesic}(p, sub_i)
  \]
  \item Population Density $ \rho $ (people/km²): Population density plays a crucial role in our model, as human factors directly influence the simulation dynamics. Areas with higher population density naturally experience greater demand for charging, leading to increased activity and movement within the region. By incorporating population density into our model, we aim to ensure that charging stations are strategically located to meet demand, optimize accessibility, and improve overall system performance.
  The population per unit area at $p$, extracted from a raster dataset (a grid-based representation of spatial data):
  \[
  \rho = \text{population\_density}(row, col)
  \]
  where row and col are the grid coordinates corresponding to 
$p$. High population density indicates greater potential demand for EV charging, making such areas more attractive for station placement.
\item Number of Nearby Stations $n_{\text{stations}} $:  
  This variable measures the number of existing charging stations within a circular region of radius $r$ centred at point $p$. By using this setting, the agent can assess the density of charging stations in its vicinity, gaining insight into the level of saturation or competition in the area. A higher concentration of stations may indicate a competitive market, where additional stations could lead to redundancy, while a lower density suggests potential opportunities for new installations. This information is crucial for optimizing station placement, balancing supply and demand, and ensuring efficient resource distribution within the network.
 \item Average Waiting Time $t_{\text{avg}}$ (hours):  
 This serves as a key indicator of local demand, highlighting regions where existing infrastructure struggles to accommodate the charging needs of users. By analyzing this measure, we can pinpoint areas requiring additional charging stations to alleviate congestion, improve service efficiency, and enhance user experience:
\end{itemize}

The complete state vector is:
\[
s = (d_{\text{nearest}},d_{\text{nearest\_sub}}, \rho, n_{\text{stations}}, t_{\text{avg}})
\]

\subsubsection{Action}

\begin{figure*}[!ht]
    \centering
    \includegraphics[width=0.8\linewidth]{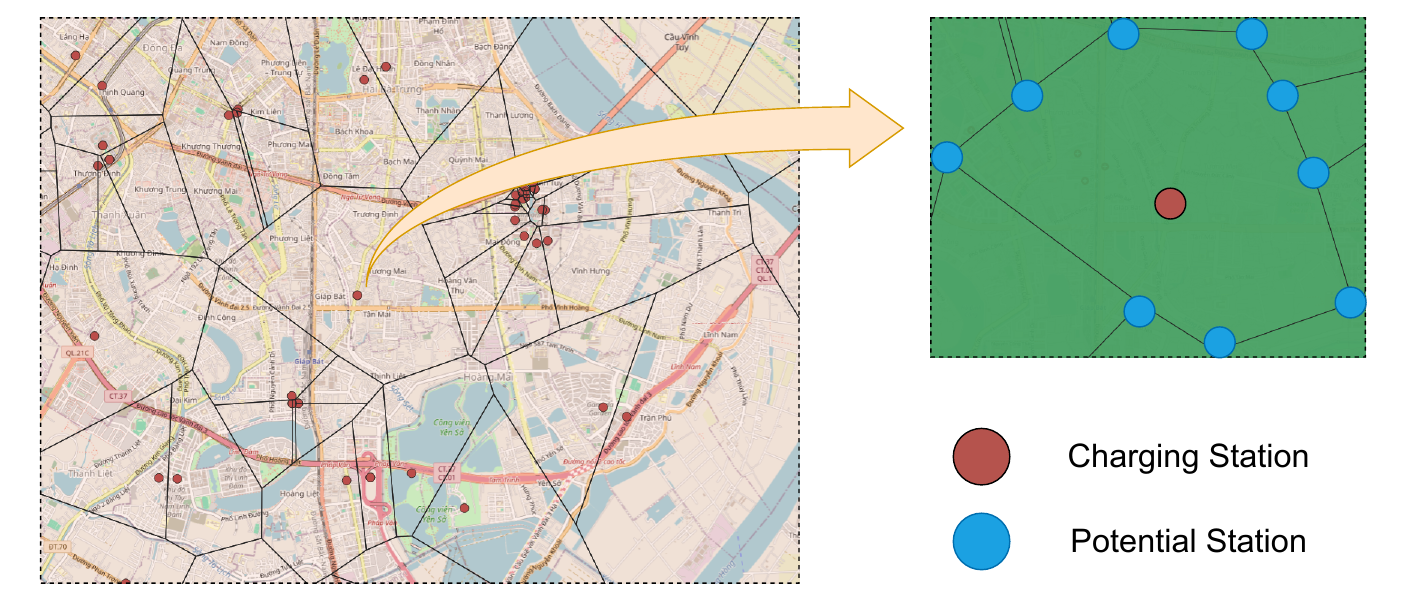}
    \caption{Voronoi region contains the vertices to make the candidate location for the charging station.}
    \label{fig:voronoi}
\end{figure*}
Action construction is a crucial step in reinforcement learning (RL) as it defines the strategy for selecting new charging station locations. The choice between discrete and continuous action spaces significantly impacts the flexibility and precision of station placement.

In earlier work \cite{padmanabhan2021optimal}, a discrete action space was employed, where the agent could move on a grid in one of four directions: left, right, up, or down. This approach simplifies decision-making but limits the granularity of station placement. More recently, a continuous action space was proposed \cite{HEO2024105567}, allowing the agent to predict precise row and column coordinates on the grid. This refinement enables finer control over station placement, leading to better alignment with real-world demand patterns.

The problem of station placement is inherently tied to Points of Interest (POIs), which act as proxies for demand distribution. Locations such as densely populated areas, commercial hubs, and high-traffic zones (e.g., highways and urban centers) are prime candidates for new charging stations. A previous study \cite{liu2023optimal} introduced a graph-based RL model with a discrete action space, where actions corresponded to selecting nodes within a road network. An attention mechanism was incorporated to help the agent focus on high-impact POIs by weighing their connectivity and demand, effectively addressing the complexity of spatial interactions.

Notably, POIs such as restaurants, museums, and retail stores serve as key proxies for electric vehicle (EV) user destinations, exerting a significant influence on charging demand \cite{wagner2013optimal, gopalakrishnan2016demand}. By leveraging these insights, RL-based approaches can better optimize station placement to meet user needs and reduce congestion

Our framework adopts a Voronoi diagram-based approach \cite{CALVOJURADO2024105719} to optimize the placement of new charging stations relative to existing infrastructure. The model prioritizes locations that maintain balanced spatial distribution between neighbouring stations, ensuring equitable coverage while minimizing service gaps. At each decision step, the agent executes a two-fold action $a$: First, selecting the geographic coordinates for the new charging station; Second, determining its operational capacity (type of charging ports). This dual optimization process simultaneously addressing placement and scalability and enables the framework to strategically enhance both the accessibility and efficiency of the charging network.

\textbf{Action for Charging Station Selection $a_{loc}$}

In our model, we take a structured geometric approach to define a discrete action space. This involves analyzing the positions of existing charging stations and generating Voronoi diagrams, which divide the area into regions based on proximity to each station. The vertices of these Voronoi regions, where boundaries intersect, are identified as potential locations for new charging stations, as shown in Figure \ref{fig:voronoi}. Each existing station's Voronoi region contributes a set of such vertices, forming a well-defined and strategic set of candidate locations for future station placement.

Let $S = \{s_1, s_2, ..., s_n\}$ be the set of existing charging stations, where n is the total number of existing stations. We construct an action space $\mathcal{P}$ based on the Voronoi diagram as points of interest following \cite{CALVOJURADO2024105719}. Let $V(S)$ be the Voronoi diagram of the existing charging station set $S$, where:
\begin{equation}
    V_i = \{x \in \mathcal{D} \mid d(x, s_i) \leq d(x, s_j) \quad \forall j \neq i\} 
\end{equation}
where $d(x, y)$ represents the Euclidean distance between points $x$ and $y$ and $V(S) = \bigcup V_i $

The action space $\mathcal{P}_i$ for each station is constructed as:
\begin{equation}
    \mathcal{P}_i = \{x \in \mathcal{D} \mid x \in \mathcal{E}(V_i)\}
\end{equation}
Here, $\mathcal{E}$ is the set of vertex all of the region $V_i$ 

The Voronoi-based solution space effectively reduces the continuous optimization problem to a discrete set of candidate points, while maintaining the geometric properties that are likely to yield optimal solutions.

\textbf{Action for Port Configuration $a_{port}$ }
\begin{table*}[!ht]
\centering
\begin{tabular}{|l|c|c|c|c|c|c|c|c|c|c|}
\hline
\textbf{Port index $j$} & 1 & 2 & 3 & 4 & 5 & 6 & 7 & 8 & 9 & 10 \\ \hline
\textbf{Power capacity (kW)} & 7 & 11 & 22 & 30 & 60 & 80 & 120 & 150 & 180 & 250 \\ \hline
\textbf{Quantity $pt_j$}        & 2 & 20 & 8  & 6  & 18 & 11 & 4   & 4   & 10  & 1   \\ \hline
\end{tabular}
\caption{Number of ports for each type of charging on the system. The variable $pt_j$ represents the quantity of charging ports for port type $j$ ($j = 1,2,...,10$).}
\label{tab: quantity}
\end{table*}
In addition to determining the placement of charging stations, we also predict their scale by forecasting the types of charging ports at each new station. Based on real-world data, we obtain the average number of each type of charging port in Table \ref{tab: quantity}. Using this information, we employ an agent to predict the probability of each port type appearing at a given station, enabling better planning and optimization of smart charging infrastructure.

The full action is:
\[
a = (a_{\text{loc}}, a_{\text{port}})
\]
\subsubsection{Reward}
The reward $r$ quantifies the quality of an action $a$ in state $s$, guiding the agent toward placements that balance population coverage, infrastructure distribution, power availability, and user convenience. It is a weighted sum of four components, which are defined below, each addressing a specific objective. Specifically, our work presents two ways to construct the reward. First, we formulate a deterministic reward that we compute directly from the static data environment:
\begin{itemize}
    \item Population Density Reward $r_{pop}$:
    Encourages stations in high-demand areas by scaling the normalized population density:
    \begin{equation}
  r_{\text{pop}} = \left( \frac{\rho}{\rho_{\text{max}}} \right) \times W_{pop}
     \end{equation}
  where $\rho_{\text{max}}$ is the maximum density in the dataset and $W_{pop}=10$ is the weight to scale the reward to a reasonable range (0 to 10). This ensures that densely populated urban areas are prioritized.
  \item Distance to Existing Stations Reward $r_{exist}$:
    \begin{equation}
  r_{\text{exist}} = \max(0, W_{exist} - d_{\text{nearest}} - d_{min})
     \end{equation}
  where $W_{exist} = 10$ is the maximum reward, and $d_{min}=1$ km is a minimum threshold below which clustering is penalized. If $d_{nearest}<1$ km, $r_{exist}=0$, discouraging overcrowding.
  \item Substation Proximity Reward $r_{sub}$:
  Ensures stations are placed near electrical substations for power supply feasibility:
    \begin{equation}
  r_{\text{sub}} = \max(0, 10 - d_{\text{sub}})
  \end{equation}
  where \( d_{\text{sub}} \) is the distance to the nearest substation, and $W_{sub}=10$. The reward decreases linearly as $d_{sub}$ increases, becoming 0 beyond 10 km.
\end{itemize}

Second, we obtain the simulation reward from the waiting time extracted from the simulation system.\\
Waiting Time Reward $r_{wati}$:
  Reduces congestion by rewarding placements that alleviate high waiting times:
\begin{equation}
r_{\text{wait}}= 
\begin{cases}
\min\left( W_{\text{wait}}, \dfrac{\tau}{t_{\text{avg}}} \right) & \text{if } t_{\text{avg}} > 0;
\\ W_{\text{wait}} & \text{otherwise}
\end{cases}
\end{equation}
where $W_{wait}=10$ is the maximum reward, and $\tau=10$ is a scaling factor. The inverse relationship with $t_{avg}$ prioritizes areas with longer waits, capping at 10 when waits are low or zero.

The total reward is:
\begin{equation}
r = r_{\text{pop}} + r_{\text{exist}} + r_{\text{sub}} + r_{\text{wait}}
\end{equation}
This reward ranges from 0 to 40, offering a well-defined signal to guide the agent's learning. In this setting, each component is assigned an equal weight, ensuring a balanced contribution from population demand, existing infrastructure, substation proximity, and waiting time reduction. 
\subsubsection{Learning for Charging Station Placement Agent}
\begin{table*}[htbp]
\centering
\label{tab:rl_comparison}
\begin{tabularx}{\textwidth}{lXX}
\toprule
\textbf{RL Algorithm} & \textbf{Advantages} & \textbf{Shortcomings} \\
\midrule
\textbf{Deep Q-Network (DQN) \cite{mnih2015human}} & 
Perfect Fit for our Action Space: Directly compatible with our discrete, Voronoi-based method for selecting station locations. & 
Outdated Method: Considered less advanced. \\
\addlinespace 

\textbf{Deep Deterministic Policy Gradient (DDPG) \cite{lillicrap2015continuous}} & 
High-Precision Placement: Can predict exact coordinates, offering finer control if the action space were continuous. & 
Incompatible by Design: Fundamentally designed for continuous action spaces, making it unsuitable for our current discrete selection method. \\
\addlinespace

\textbf{Proximal Policy Optimization (PPO) \cite{schulman2017proximal}} & 
Versatile \& Modern: Works with our existing discrete action space while being a more stable, state-of-the-art algorithm. & 
Lower Sample Efficiency: As an on-policy method, it may require more environmental interactions to learn compared to an off-policy method like DQN. \\
\bottomrule
\end{tabularx}
\caption{Concise Comparison of RL Algorithms for EVCS Placement}
\label{table: comparision}
\end{table*}
We present the advantages and shortcomings of three RL algorithms in table \ref{table: comparision}. From our perspective, DQN and PPO are suitable for our discrete action space, which is particularly well-suited for handling Voronoi-based and port configuration action spaces. While DDPG is better suited for purely continuous action spaces, which might not fit our Voronoi-based location selection.
From this comparison, we select the Deep Q-Network (DQN) as the agent, since our dataset is relatively simple. Furthermore, training DQN is more sample-efficient than PPO, making it a more suitable choice in our setting. In addition, DQN’s value-based framework is well aligned with discrete action spaces, which further supports its effectiveness for our problem. Since our framework also integrates simulation, which is computationally expensive, the higher sample efficiency of DQN is particularly advantageous.

To optimize both charging station selection and port configuration, the agent employs a hybrid action space handled by two deep Q-networks. The learning process involves parameter updates based on environmental feedback, with separate networks for location decisions ($a_{\text{loc}}$) and port configuration ($a_{\text{port}}$).

We employ two decoupled networks to address the combinatorial action space:
\begin{itemize}
    \item Location Q-network ($Q_\theta$): Selects optimal charging station location.
    \item Port Configuration Q-network ($Q_\phi$): Chooses port configuration, which is the type of port associated with the quantity. 
\end{itemize}

In DQN, we have the online network and the target network, which are two neural networks with identical architectures but distinct roles. The online network directly interacts with the environment to select actions and continuously updates the parameters via gradient descent during training. The target network is a separate network that's a periodic copy of the online network. It provides stable Q-value targets for training the online network. 

\textbf{Location Selection Q-Network}
The location Q-network updates its parameters $\theta$ through temporal difference learning:
\begin{equation}
Q_{\theta}(s, a_{\text{loc}}) \leftarrow Q_{\theta}(s, a_{\text{loc}}) + \Delta Q
\end{equation}
where
\begin{equation*}
\Delta Q = \alpha \left( r + \gamma \max_{a'} Q_{\theta^-}(s', a') - Q_{\theta}(s, a_{\text{loc}}) \right)
\end{equation*}

Here, $\theta$ is the online network parameters, $\theta^-$ is the target network parameters (updated periodically), $\alpha$ is the learning rate, $\gamma$ is the discount factor.

The corresponding loss function minimizes temporal difference error:
\begin{equation}
L_{\text{loc}}(\theta) = \mathbb{E}_{(s, a, r, s') \sim D} \left[ \left( Q_{\theta}(s, a_{\text{loc}}) - y_{\text{loc}} \right)^2 \right]
\end{equation}
where target values $y_{\text{loc}}$ are computed as:
\begin{equation}
y_{\text{loc}} = r_{\text{loc}} + \gamma \max_{a'} Q_{\theta^-}(s', a')
\end{equation}
To predict the next action, we choose the action that maximizes the Q value: $a_{loc}=\arg \max_a Q_\theta (s, a)$

\textbf{Port Configuration Q-Network}
The port network $\phi$ handles technical parameters through constrained Q-learning:
\begin{equation}
Q_{\phi}(s, a_{\text{port}}) \leftarrow Q_{\phi}(s, a_{\text{port}})  + \Delta Q
\end{equation}
where $\Delta Q=
\beta \left( r + \gamma\max_{a''} Q_{\phi}(s', a'')) - Q_{\phi}(s, a_{\text{port}}) \right)$

where $\phi$ is the online network parameters, $\phi^-$ is the parameter of the target network, and $\beta$ is the port-specific learning rate,

Port configuration loss with double Q-learning stabilization:
\begin{equation}
L_{\text{port}}(\phi) = \mathbb{E}_{(s, a, r, s') \sim D} \left[ \left( Q_{\phi}(s, a_{\text{port}}) - y_{\text{port}} \right)^2 \right]
\end{equation}
\begin{equation}
y_{\text{port}} = r_{\text{port}} + \gamma \max_{a''} Q_{\phi}(s', a''))
\end{equation}


\subsection{Simulation-based Gama model}
We utilize the Gama\footnote{http://gama-platform.org} \cite{Taillandier2019} simulation model to replicate the movement of EVs within a city. This simulation tracks the vehicles as they navigate the city and search for available charging stations. By analyzing this behavior, we extract key metrics such as the average waiting time at each charging station, providing valuable insights into station usage and efficiency.
\begin{figure*}[!h]
    \centering
    \includegraphics[scale=0.75]{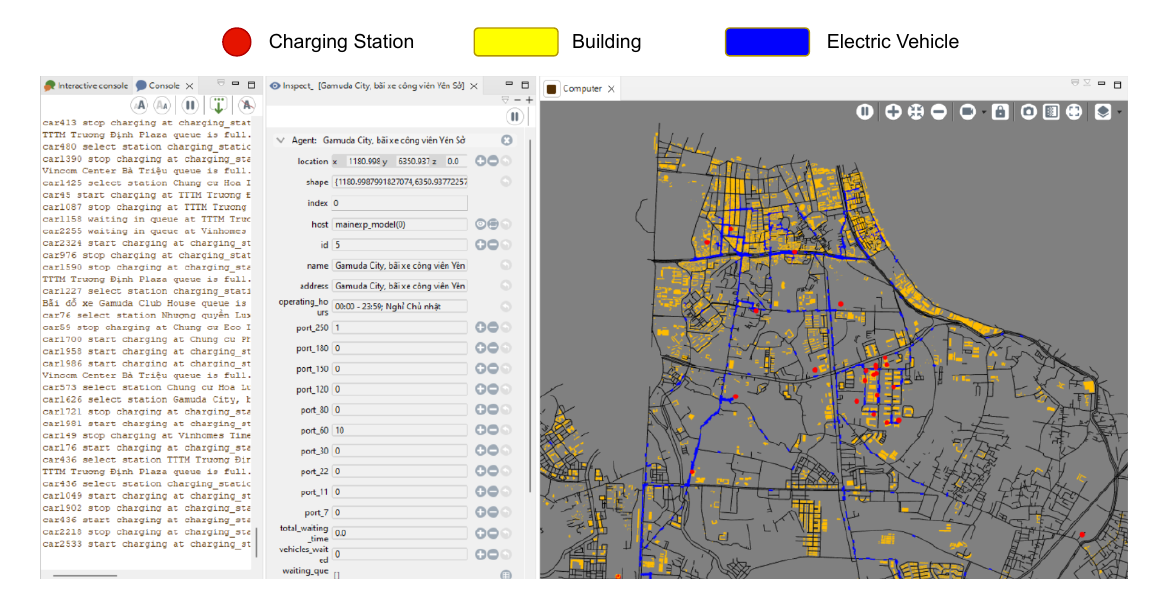}
    \caption{Simulate the process of moving and searching for a station to charge the electric vehicle in GAMA.}
    \label{fig:simulation}
\end{figure*}
\begin{figure}[!h]
    \centering
    \includegraphics[width = 0.9\linewidth]{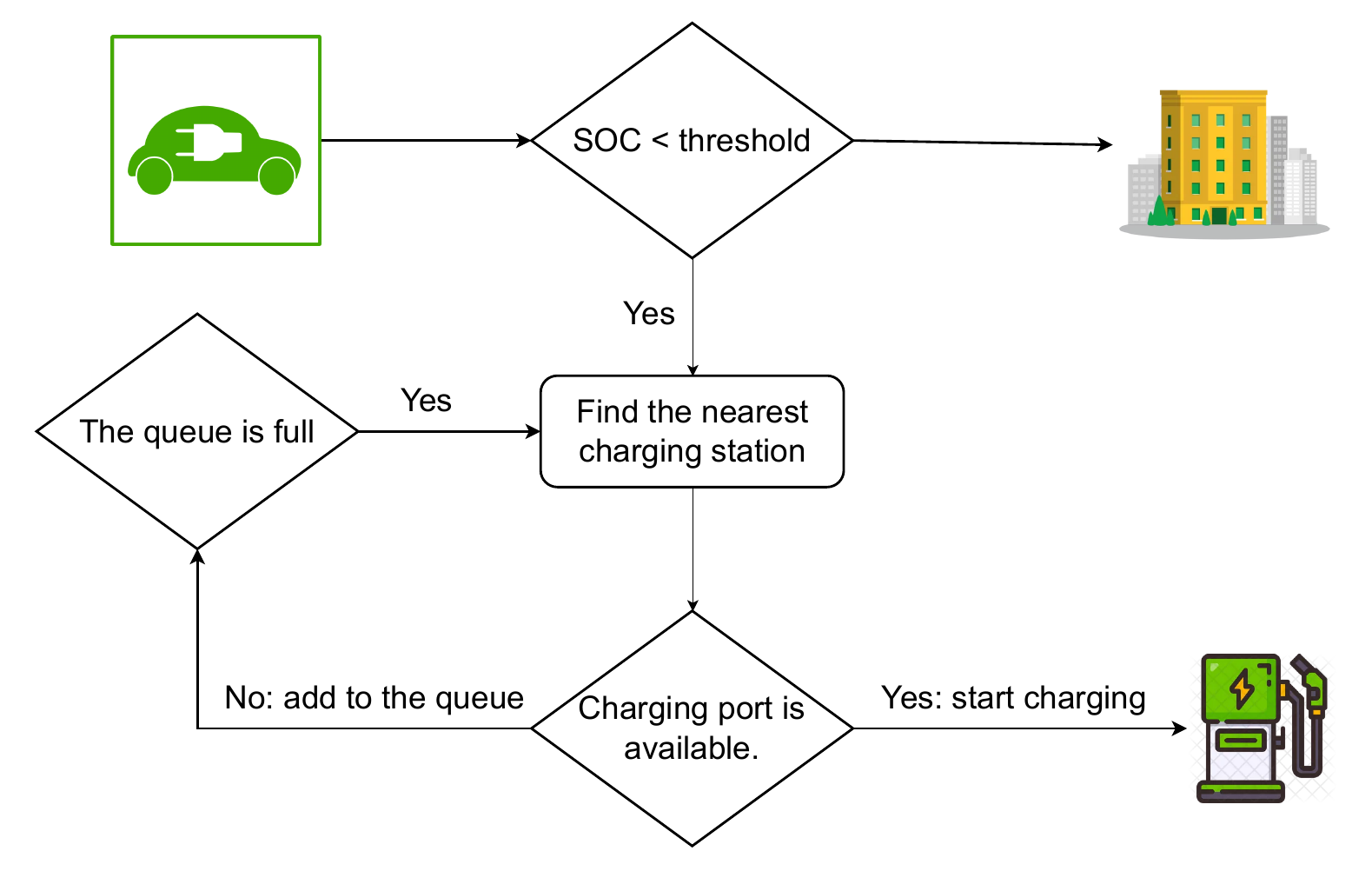}
    \caption{Flowchart of Charging Behavior Based on Battery Level and Station Availability}
    \label{fig: simulation diagram}
\end{figure}
The simulation consists of three main components:
\begin{itemize}
    \item Vehicle Agents: Each EV is represented as an autonomous agent with decision-making capabilities based on battery state-of-charge (SoC). Due to the absence of user behavior data, we simulated random user behavior based on online research. Specifically, when the SoC reaches 30\%, users randomly decide whether to start charging. Similarly, when the SoC reaches 80\% during charging, users determine whether to stop or continue charging.  
    \item Charging Stations: Charging stations have a limited number of charging points and operate under various queueing policies, including First-Come, First-Served (FCFS). By default, when an EV arrives at a charging station, it is assigned the largest available charging port. To better reflect real-world conditions, we impose queue length constraints at each station based on its capacity and scale.
    
    \item Environment: The simulation environment encompasses key elements such as road networks, charging station locations, traffic conditions, and surrounding infrastructure, including buildings. These factors collectively influence the movement and behavior of EVs within the system.
\end{itemize}

In general, vehicle agents navigate the road network, traveling between different buildings. To enhance realism, target buildings are randomly selected \cite{JORDAN2022116739}, ensuring that the distance to the destination exceeds a predefined minimum threshold. When the battery level drops to a predefined threshold SOC, the vehicle searches for the nearest charging station. If a charging slot is available, the vehicle begins charging; otherwise, it joins a waiting queue. A maximum limit is set for the queue, and if it is full, the vehicle will seek the next closest charging station following in Figure \ref{fig: simulation diagram}.

During movement, each vehicle is assigned a random speed within the range of 40–50 km/h, ensuring suitability for the traffic density in our case study. Each station is equipped with a mix of fast and slow charging ports. To simplify the process, fast charging is always prioritized when available.


In our simulation system, an important advantage is its ability to adapt to the increasing number of EVs expected in the near future. As EV adoption grows, the demand for charging will experience a significant surge, potentially leading to congestion at charging stations. By simulating large-scale EV deployment, we can assess the impact of infrastructure expansion, alternative energy sources, and smart charging policies, ensuring the system remains efficient and scalable as demand continues to rise.

\begin{algorithm}[!h]
    \caption{Station Placement Algorithm}
    \begin{algorithmic}[1]
        \State \textbf{Initialization}: Define the number of stations and episodes.
        
        \For{each episode}
            \State Initialize with a single random station.
            \For{each step in the episode}
                \State \textbf{Voronoi Region Generation:} Compute Voronoi regions based on current station locations and extract vertices as candidate locations.
                
                \State \textbf{State Extraction:} Compute the state vector $s$ for the current station configuration.
                
                \State \textbf{Action Selection:}
                \State Select station location $a_{\text{loc}}$ using the Location Selection Q-Network.
                \State Select port configuration $a_{\text{port}}$ using the Port Configuration Q-Network.
                
                \State \textbf{Environment Update:}
                \State Place a new station at $a_{\text{loc}}$ with ports configured as $a_{\text{port}}$.
                \State Update the environment (recompute Voronoi regions, configure ports).
                \State Simulate the system to compute reward $r$ and observe next state $s'$.
                
                \State \textbf{Experience Storage:} Store the tuple $(s, a, r, s')$ in the replay buffer.
                
                \State \textbf{Learning:}
                \State Sample a mini-batch from the replay buffer.
                \State  Alternating updates: Location and port networks trained.
    \State Target network synchronization: $\theta^- \leftarrow \tau\theta + (1-\tau)\theta^-$, 
     \\$\phi^- \leftarrow \tau\phi + (1-\tau)\phi^-$ (soft update)
                
                \State \textbf{State Transition:} Assign the newly placed charging station as the current station.
            \EndFor
        \EndFor
    \end{algorithmic}
\end{algorithm}

\section{Experiments}
\subsection{Data Collection}
To demonstrate the effectiveness of our approach, we collect data about the population density \cite{CIESIN2017} from the Center for International Earth Science Information Network - CIESIN - Columbia University. They measured the number of persons per square kilometer, derived from national census data and population registers for the years 2000, 2005, 2010, 2015, and 2020.

The data for the existing charging stations, electric vehicles, and charging sessions at these stations are collected from a private database owned by a private company. This data, which includes detailed information on the charging sessions such as time, duration, and energy consumption, is not publicly available and will not be published. The data for road networks and buildings are sourced from OpenStreetMap\footnote{https://wiki.openstreetmap.org/wiki/Map\_features}. We use QGIS\footnote{https://qgis.org/} software to extract, process, and analyze this data efficiently.
\subsection{Experiments Setting}
\subsubsection{Simulation Settings.} To begin with, we initialized a fleet of 3000 electric vehicles, each with varying states of charge (SoC), to simulate their movement across two districts in Hanoi, Vietnam. The simulation area is equipped with 30 existing charging stations distributed throughout the region. For this simulation, we selected a range of modern electric vehicles currently available in Vietnam, including the VFe34, VF8, and VF9 models \footnote{https://vinfast.com/}, which have battery capacities of 42 kWh, 87.7 kWh, and 123 kWh, respectively. These vehicles were chosen to represent a broad spectrum of electric vehicle sizes and capabilities, reflecting the diversity of the Vietnamese market. To evaluate the performance of the charging infrastructure, we conducted a 24-hour simulation of the system. This simulation aims to capture key metrics such as the waiting time for charging and the overall charging time at each station.

\subsubsection{Hyperparameter for Deep Q-network }
\begin{table}[!ht]
\centering
\resizebox{\linewidth}{!}{ 
\begin{tabular}{lccc}
\toprule
\textbf{Layer} & \textbf{Input Size} & \textbf{Output Size} & \textbf{Activation Function} \\
\midrule
Input Layer & --- & \textit{state\_size} & --- \\
\addlinespace
Fully Connected 1 (fc1) & \textit{state\_size} & 256 & ReLU \\
Fully Connected 2 (fc2) & 256 & 128 & ReLU \\
Fully Connected 3 (fc3) & 128 & 64 & ReLU \\
\addlinespace
Output Layer (fc4) & 64 & \textit{action\_size} & None (Linear) \\
\bottomrule
\end{tabular} }
\caption{Deep Q-Network Architecture}
\label{tab:q_network_architecture}
\end{table}
For the deep Q-network (DQN) architecture, we designed a network consisting of three fully-connected layers. Each layer is fully connected to the subsequent layer, with a Relu activation function applied after each hidden layer to introduce non-linearity and enhance the network's learning capabilities (Table \ref{tab:q_network_architecture}). The DQN is trained over the course of 100 episodes, with the agent taking 10 steps per episode. 
For optimizing the learning process, we set the following hyperparameters for the entire optimization problem: the discount factor $\gamma$ is set to 0.99, which helps balance immediate and future rewards; the exploration rate $\epsilon$ is initialized at 1.0 to encourage exploration during the early stages of training and is decayed over time to shift towards exploitation of the learned policy; the batch size is set to 8 to ensure that each update is based on a diverse set of experiences, providing stable and reliable learning; and the learning rate is set to 0.001 to control the magnitude of updates to the Q-values, ensuring gradual convergence.
\subsection{Baseline Method}
To evaluate the effectiveness of our framework, we perform a comparative analysis against a greedy baseline algorithm, referred to as the Voronoi-based method \cite{CALVOJURADO2024105719}. We automatically generate charging station locations using two different strategies: radial and probabilistic distributions \cite{marti2020load, jordan2022electric}. For the probabilistic distribution, population density is used as a guiding factor areas with higher population density have a higher probability of being selected as potential charging station sites. This ensures that the charging infrastructure is more likely to be placed where demand is expected to be higher.

The baseline models do not account for the variation in charging station scales. To address this, we conducted experiments using two strategies. First, we randomly assigned different types of charging ports to each station, with the assignment probabilities based on the observed frequency distribution of port types in our existing real-world system. Second, we selected the port type based on a probability distribution influenced by the local population density.
\subsection{Evaluation Metrics}
The waiting time $(wait)$ represents the mean waiting time experienced by all vehicles across all charging stations. A lower waiting time indicates a more efficient distribution of charging resources and reduced congestion at charging points.

The charging time $(charging)$ captures the total amount of time spent charging within the entire simulation period. Lower charging times reflect an optimized station network where vehicles can access and complete charging more quickly, minimizing overall system load.
\section{Results}
\subsection{Simulation Results}
\begin{figure}[!t]
    \centering
    \begin{subfigure}{0.5\textwidth}
        \centering
        \includegraphics[width=\linewidth]{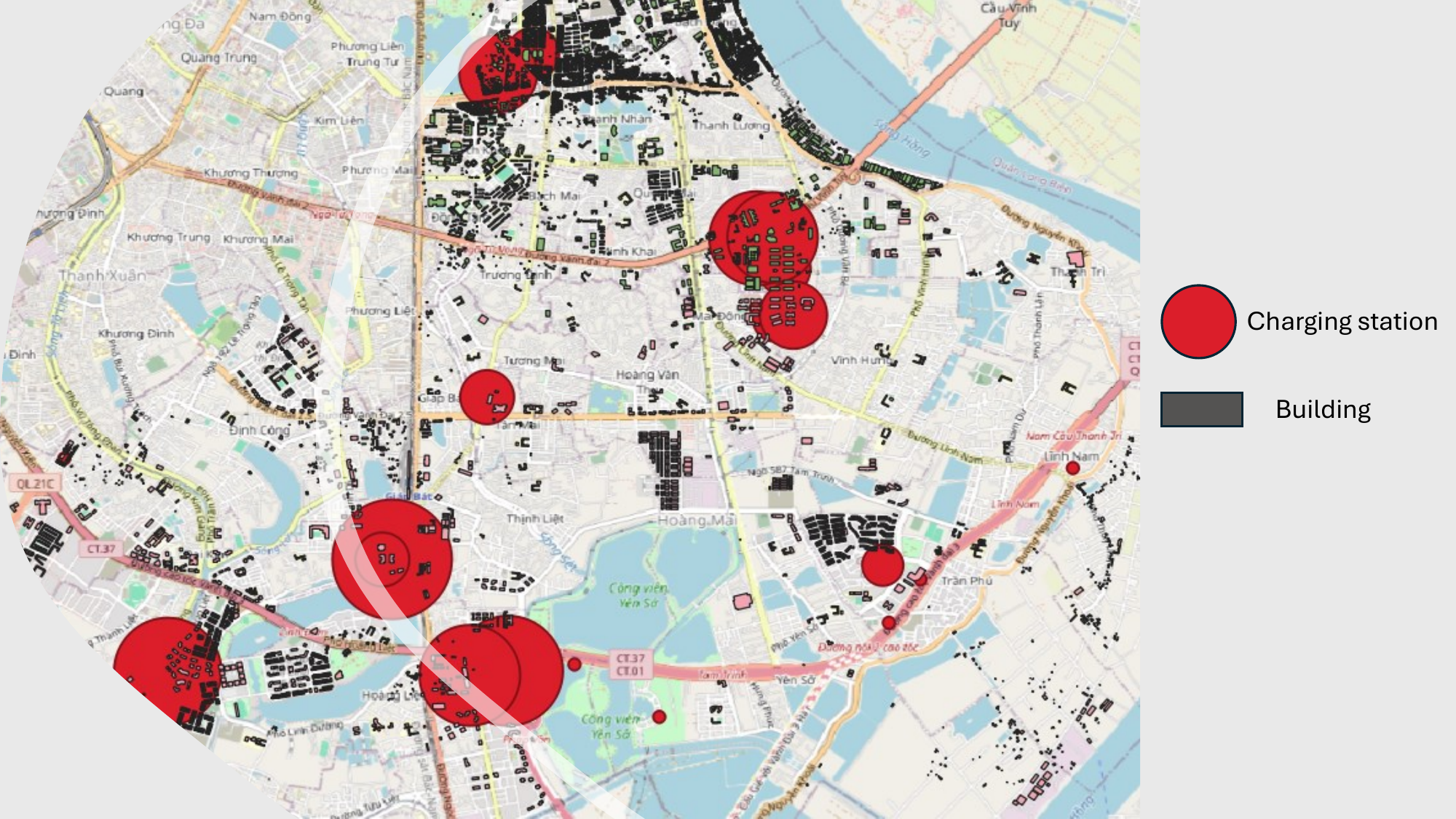}
        \caption{Current Waiting Time Trends at Charging Stations}
        \label{fig:simulation-sub1}
    \end{subfigure}
    \hfill
    \begin{subfigure}{0.35\textwidth}
        \centering
        \includegraphics[width=\linewidth]{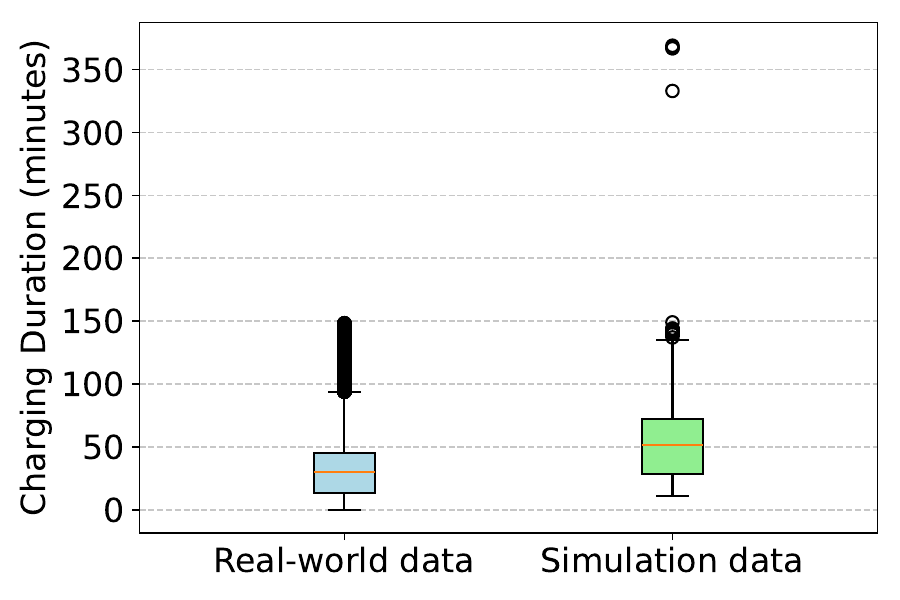}
        \caption{Boxplot of charging times in simulation vs. real-world data.}
        \label{fig:simulation-sub2}
    \end{subfigure}
    \caption{Simulation results}
    \label{fig:simulation result}
\end{figure}

In figure \ref{fig:simulation result}, we present the waiting times for all stations. Based on our simulation, we extract the waiting time for each station, which is represented in the form of circles. The size of each circle corresponds to the waiting time: larger circles indicate longer waiting times. Our observations reveal that stations with longer waiting times tend to be located in areas with a high density of large buildings. This issue becomes particularly evident in areas with high population density, such as workplaces or public buildings. The high volume of vehicle movement in these locations leads to a significant demand for charging. Current status indicates that the existing charging system is insufficient to meet this demand.
\begin{figure*}[!ht]
    \centering
    \begin{subfigure}{0.45\textwidth}
        \centering
        \includegraphics[width=\linewidth]{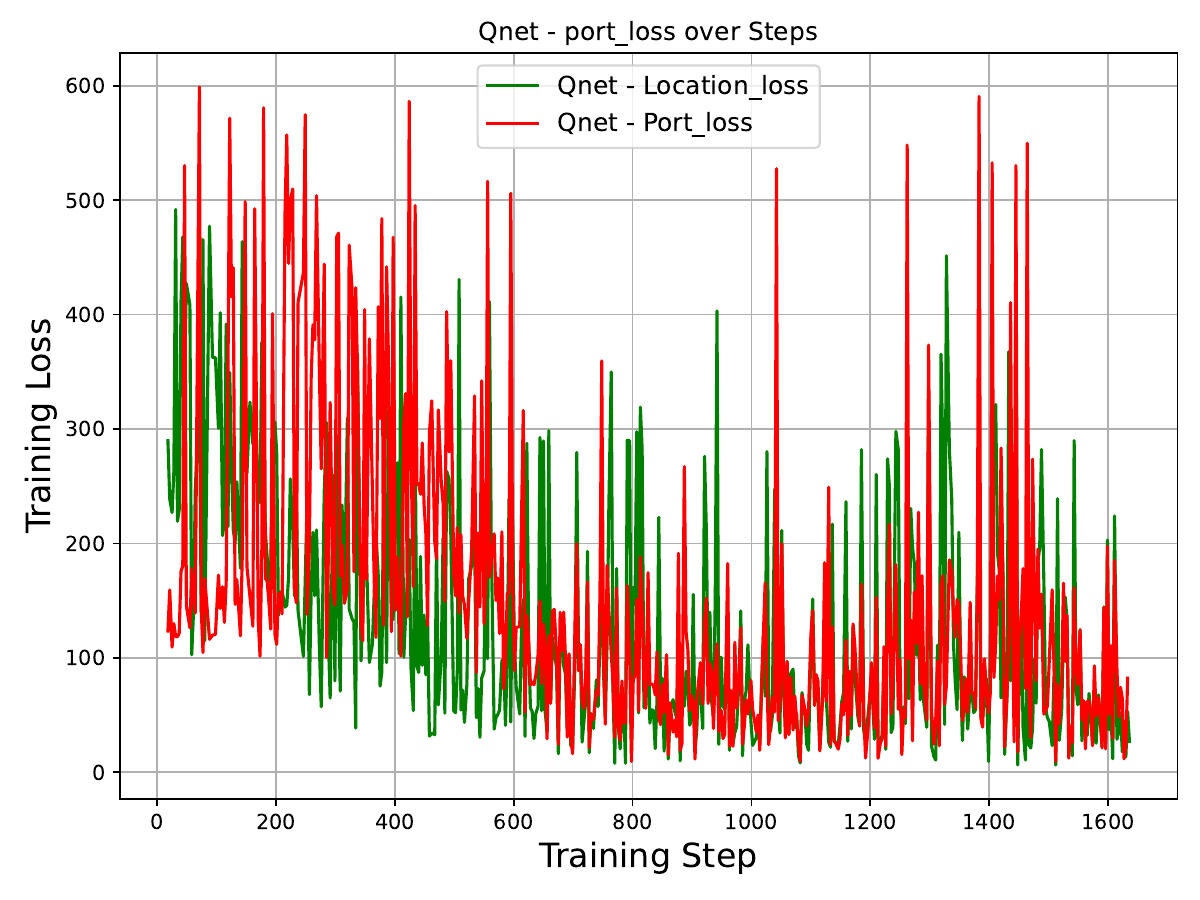}
        \caption{Loss value of the location and port configuration Q-network}
        \label{fig:sub1}
    \end{subfigure}
    \hfill
    \begin{subfigure}{0.45\textwidth}
        \centering
        \includegraphics[width=\linewidth]{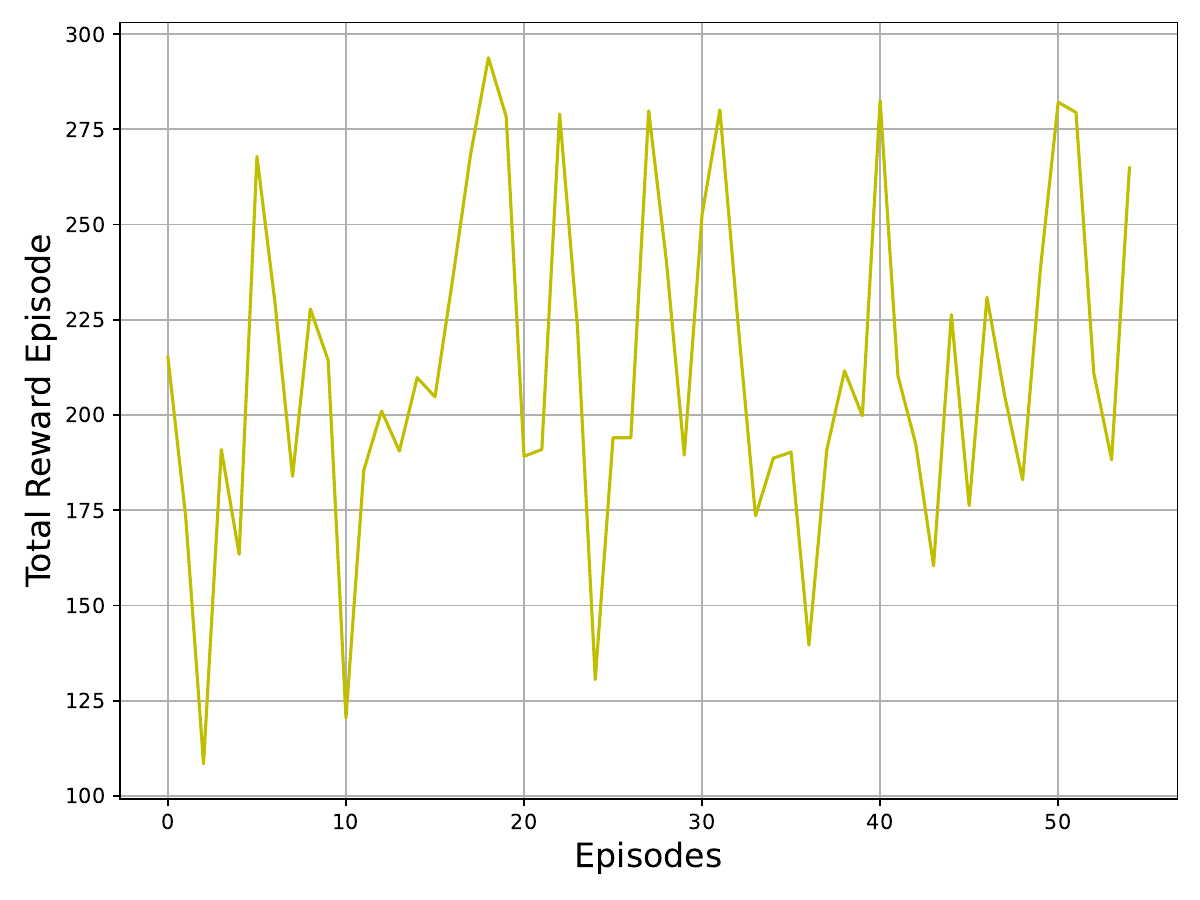}
        \caption{Total reward over episodes}
        \label{fig:sub2}
    \end{subfigure}
    \caption{Training Q-network}
    \label{fig:loss}
\end{figure*}
In figure \ref{fig:simulation-sub2}, the charging times for both systems are comparable. Our simulation framework is designed to closely mimic real-world user behavior, with an average charging time of approximately 50 minutes. Although there are some differences in the median and spread, the boxplots remain reasonably comparable, indicating that the simulation system effectively approximates real-world charging times.
\subsection{RL Training Performance}
To evaluate the learning dynamics and stability of our hybrid deep reinforcement learning (DRL) framework, we examined key training metrics, including the loss of the dual Q-networks and the reward progression across episodes. The training results highlight both the challenges and effectiveness of our hybrid DRL framework. In Fig. \ref{fig:sub1}, although the training losses of the location and port Q-networks fluctuate significantly, this observation is expected due to the stochastic nature of deep Q-learning with replay buffers and $\epsilon$-greedy exploration. Despite the oscillations, both losses exhibit a downward trend, indicating that the networks progressively approximate more accurate Q-value estimates. Moreover, this confirms that employing two Q-networks simultaneously does not hinder convergence. Instead, the dual-network design ensures that both components effectively learn their respective tasks while jointly contributing to the overall optimization process. Meanwhile, Fig. \ref{fig:sub2} shows that the total reward generally follows an upward trajectory, reflecting gradual improvement over episodes.
\subsection{Result of Charging Station Placement}
In table \ref{tab:1}, we report the results for both the baseline method and our proposed framework. For each baseline result, the left side illustrates a random port type selection, while the right side shows a selection based on population density. Our framework shows the outperforming result when we compare it to the baseline methods and our framework without incorporating the waiting time. It is clear that our approach, which uses reinforcement learning, is suitable in the context of the real environment for placing charging stations. Moreover, integrating the waiting time and aiming for flexible feedback brings a good result, which demonstrates the effectiveness of simulation reward instead of only a deterministic reward computed from the environment data. Besides, our framework without using the waiting time has a significant improvement, which demonstrates the effectiveness of the RL-based approach is more suitable and flexible when selecting the next charging station.
\begin{table}[!t]
\centering
\resizebox{0.8\linewidth}{!}{ 
\begin{tabular}{|l|c|c|c|c|}
\hline
\textbf{Method}       & \multicolumn{2}{c|}{\textbf{Wait (hours) $\downarrow$}}  & \multicolumn{2}{c|}{\textbf{Gap (\%) $\uparrow$}} \\ \hline
Original state        & \multicolumn{2}{c|}{1.7360}                & \multicolumn{2}{c|}{0}                      \\
Voronoi based         & 1.2037 & 1.1793              & 30.66 & 32.07                \\
Radial         & {\ul0.9774} & 0.9976            & {\ul43.7} & 42.53            \\
Probabilistic         & 1.5792 & {\ul0.9343}            & 9.03 & 46.18          \\
Ours w/o waiting time & \multicolumn{2}{c|}{0.9894}         & \multicolumn{2}{c|}{43.00}           \\ \hline
\textbf{Ours}         & \multicolumn{2}{c|}{\textbf{0.9251}}       & \multicolumn{2}{c|}{\textbf{53.28}}         \\ \hline
\end{tabular}
} 
\caption{Comparing the waiting time between our framework and the Voronoi-based method. } \label{tab:1}
\end{table}
\begin{figure}[!t]
    \centering
    \includegraphics[width=0.8\linewidth]{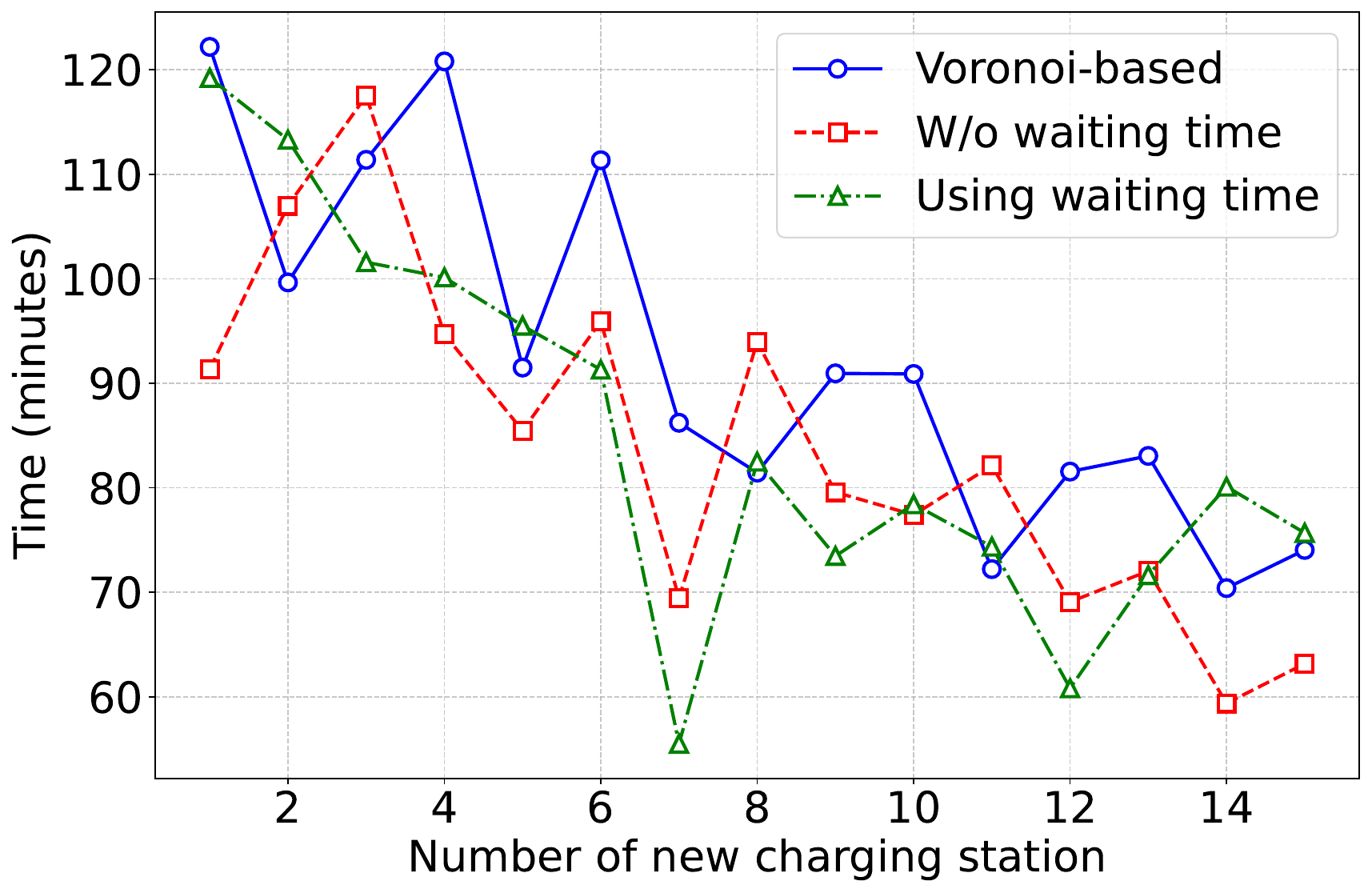}
    \caption{Waiting time reduction trends and optimal points}
    \label{fig: demand}
\end{figure}
Based on the waiting time estimated for 15 new charging stations in Figure \ref{fig: demand}, our method necessitates the addition of six new charging stations to achieve a 53\% reduction in overall system waiting time, demonstrating superiority over alternative methods. In contrast, approaches that do not account for waiting time and the number of charging ports struggle to accurately predict the required scale of charging stations, resulting in significant fluctuations in performance. Our method exhibits a steady decrease in waiting time, with occasional oscillations, but ultimately trends downward. The increase in waiting time, even with the addition of charging systems, occurs because the current scale of charging stations fails to meet demand. By integrating waiting time into the learning reward, our method enhances the prediction of the necessary charging station scale. Consequently, when waiting time increases, the next step involves placing a new charging station at an appropriate scale to effectively reduce waiting time. In this result, we do not present radial, probabilistic methods because the points are selected as a set instead of proposing each point over time.

\subsection{Ablation Study}
\subsubsection{Comparative Analysis of RL Algorithms}
\begin{figure}[!t]
    \centering
    \begin{subfigure}{0.45\textwidth}
        \centering
        \includegraphics[width=\linewidth]{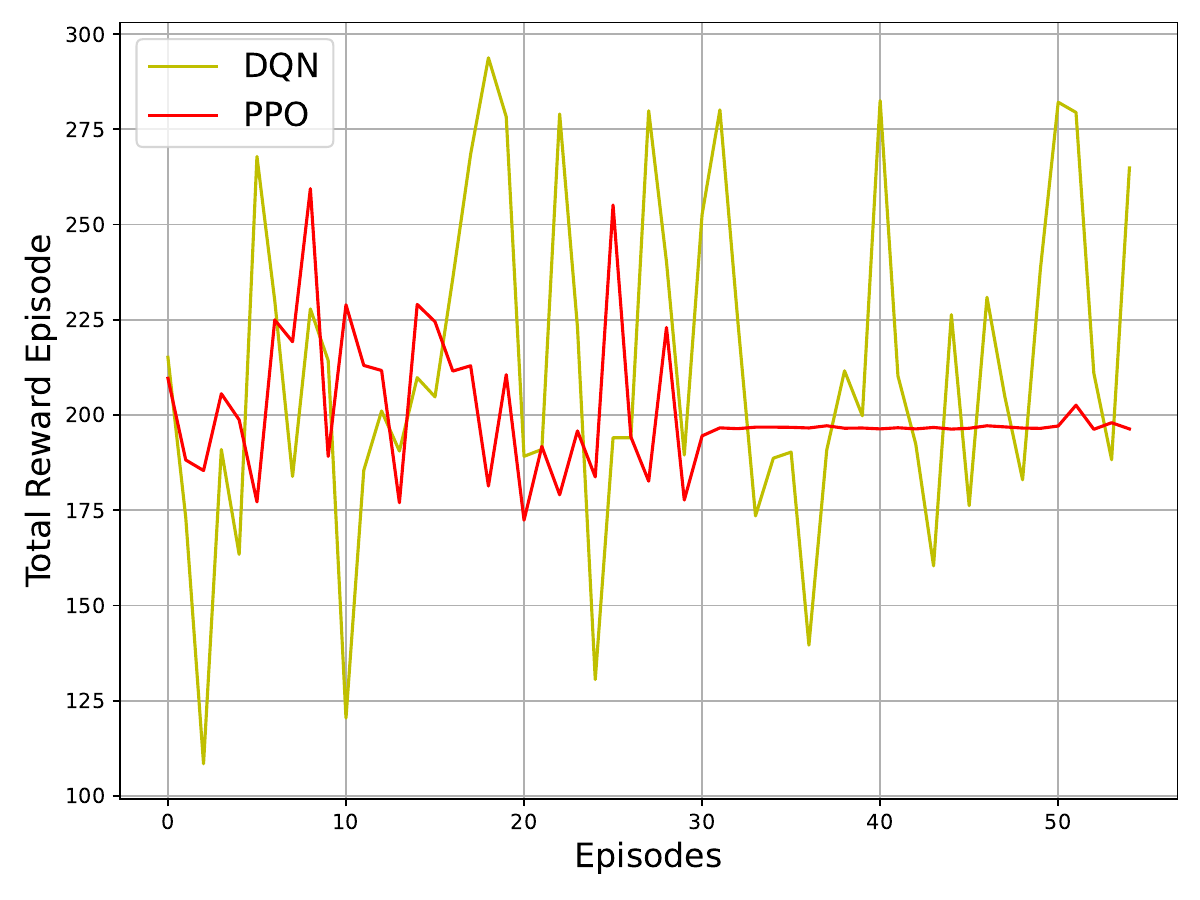}
        \caption{Total reward during training for DQN and PPO}
        \label{fig:sub3}
    \end{subfigure}
    \hfill
    \begin{subfigure}{0.45\textwidth}
        \centering
        \includegraphics[width=\linewidth]{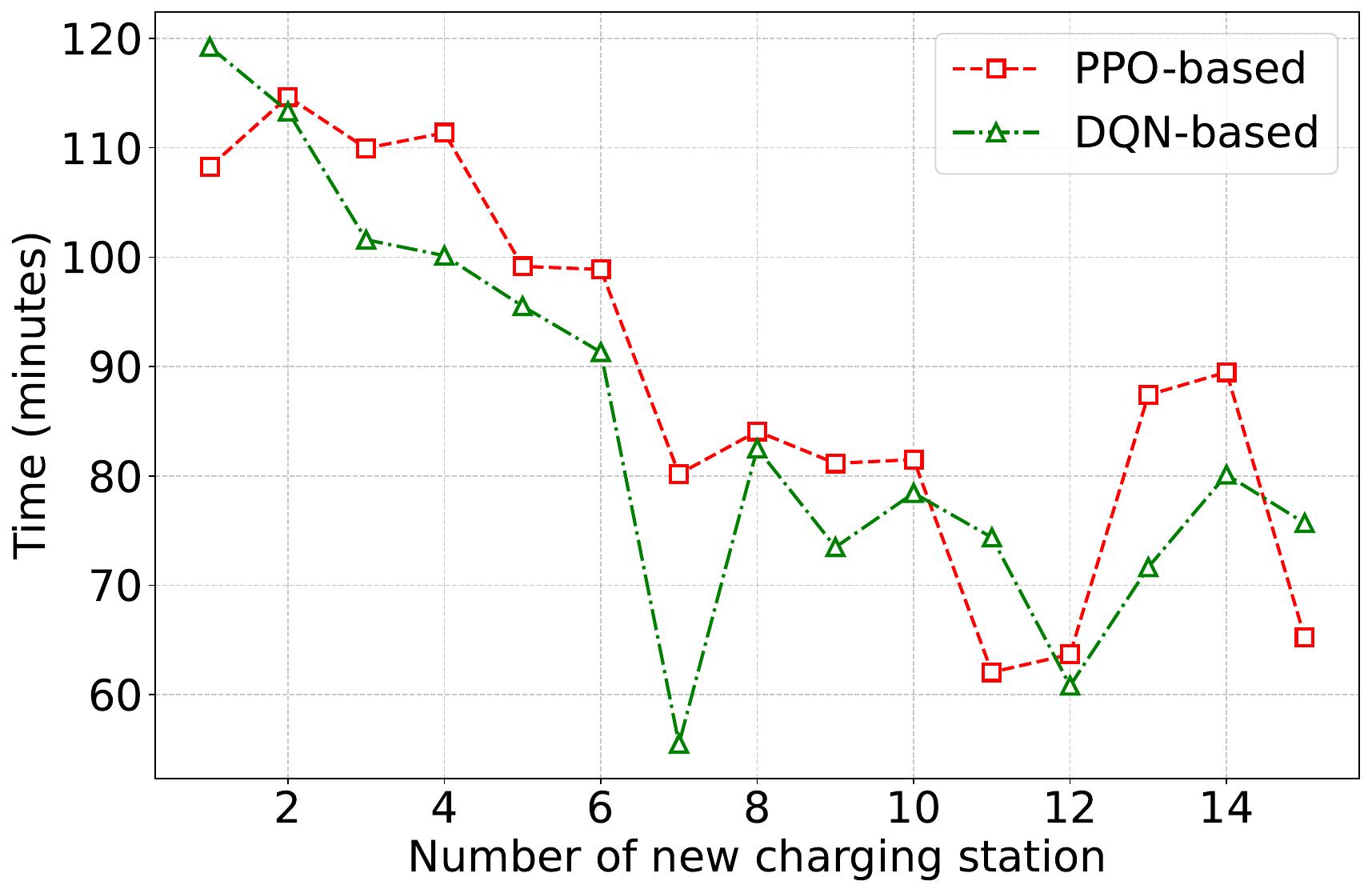}
        \caption{Waiting time trend with increasing numbers of new charging stations.}
        \label{fig: sub4}
    \end{subfigure}
    \caption{Comparison of two RL models (Deep Q-network - DQN and Proximal Policy Optimization - PPO) in our algorithm.}
    \label{fig:loss}
\end{figure}

In Fig. \ref{fig:sub3}, we present the total reward over episodes for both models. The results indicate that the PPO algorithm is not well suited to our framework, as its reward remains relatively stable without a clear upward trend. In contrast, the DQN model demonstrates higher variability but achieves a generally increasing reward, suggesting better adaptation and learning effectiveness in this setting. Fig. \ref{fig: sub4} illustrates the waiting time trend as the number of new charging stations increases. Both approaches show a clear reduction in waiting time as more stations are added, confirming that system capacity expansion directly improves user experience. However, the DQN-based method consistently achieves lower waiting times than the PPO-based approach, especially when the number of stations exceeds five. This suggests that DQN not only adapts more effectively during training but also yields more efficient station placement decisions in practice.
\subsubsection{The coverage of new charging stations}
\begin{figure*}[!h]
    \centering
    \includegraphics[width=0.75\linewidth]{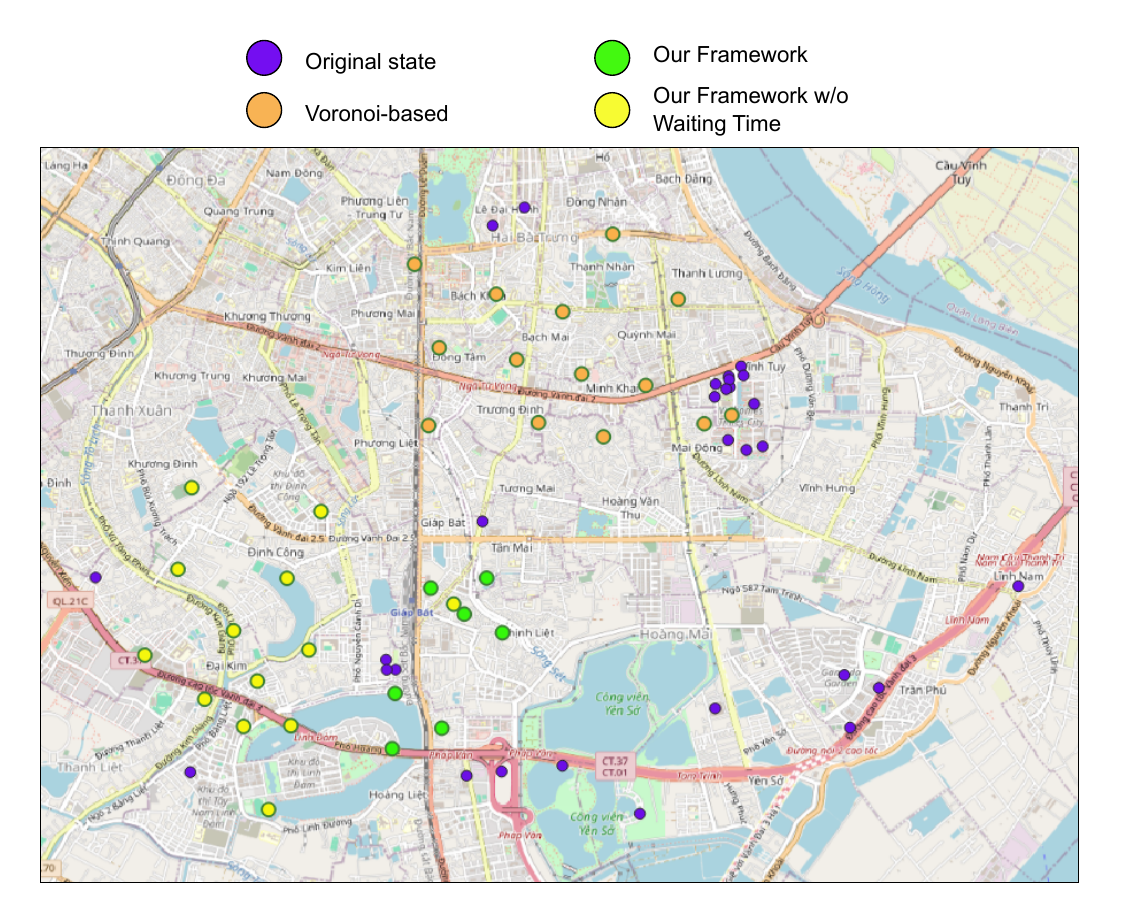}
    \caption{Distribution of charging stations between methods.}
    \label{fig: coverage}
\end{figure*}
\begin{table}[!ht]
\centering
\resizebox{0.8\linewidth}{!}{
\begin{tabular}{lc}
\hline
\textbf{Method}                & \textbf{Proximity to nearest CS (km)} \\ \hline
\textbf{Voronoi-based}         & 0.8950             \\ 
\textbf{Radial}                & 0.6503             \\ 
\textbf{Probabilistic}         & 1.4246             \\ 
\textbf{Ours w/o waiting time} & 0.9454             \\ \hline
\textbf{Ours}                  & 0.6300             \\ \hline
\end{tabular}
}
\caption{The mean distance of the new charging station proximity to the nearest existing station.}
\label{tab:distance}
\end{table}
In table \ref{tab:distance}, we present the distances of the new charging stations relative to existing stations. Our framework prioritizes selecting the next station that is closest to the existing stations for two reasons: first, a new charging station located nearby can better accommodate the high demand in these areas; second, when the queue at a station is full, users can easily switch to the next station to charge their vehicles. On the contrary, the remaining methods, although having better coverage as shown in Figure \ref{fig: coverage}, have worse performance. This again acknowledges that good coverage, but not matching the actual needs causes worse performance.

\subsubsection{Future Growth of Charging Stations}
\begin{figure}[!h]
    \centering
    \includegraphics[width=0.43\linewidth]{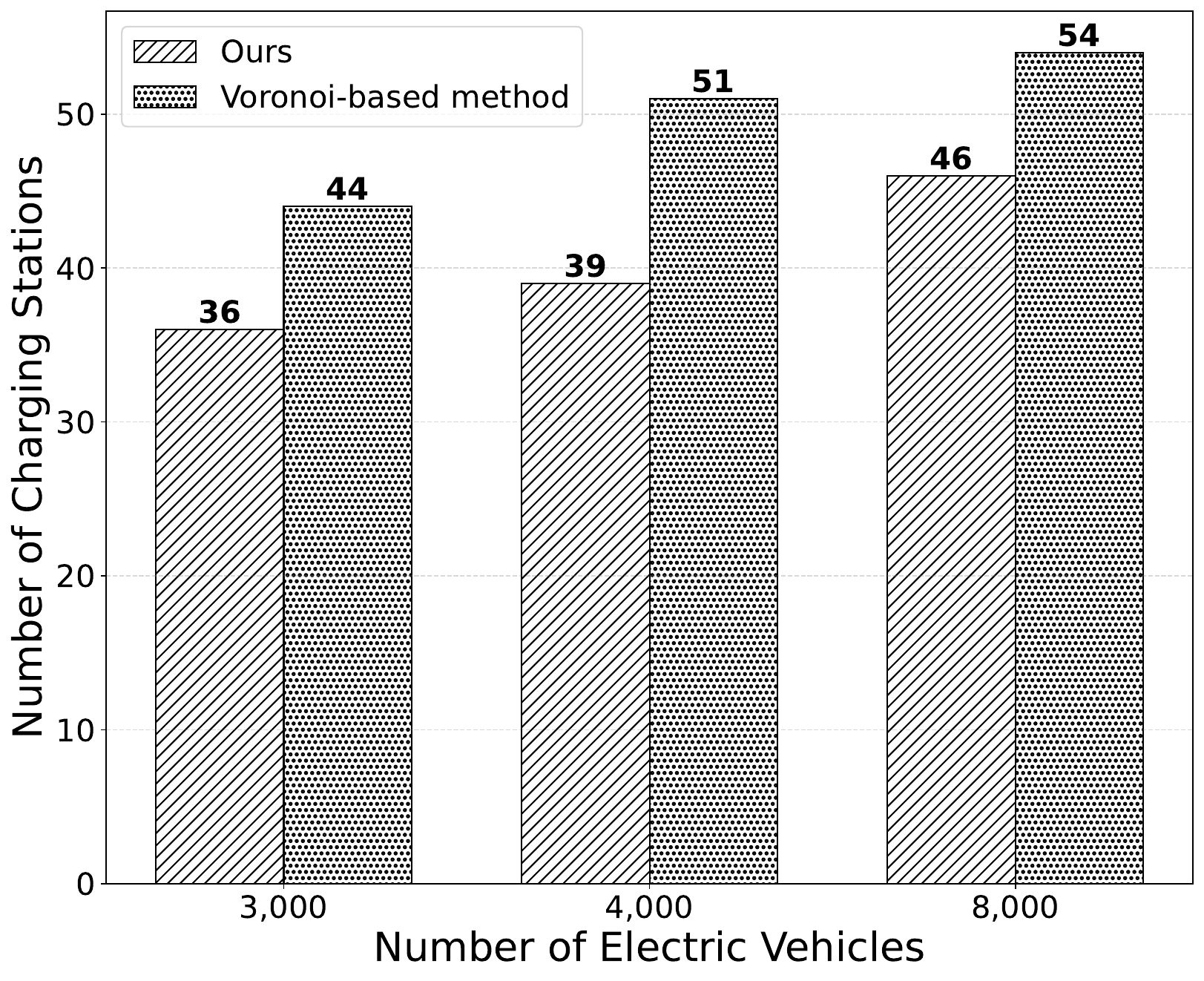}
    \caption{Optimal Charging Station Deployment under Future EV Demand.}
    \label{fig:electric_vehicles_charging_stations_comparison}
\end{figure}
Under current conditions, how many charging stations should be established to meet higher demand levels? 
In Figure \ref{fig:electric_vehicles_charging_stations_comparison}, we present a comprehensive analysis of the optimal number of charging stations needed to accommodate future increases in electric vehicle demand. The performance of our approach demonstrates a significant advantage, particularly when the number of charging stations is lower than that proposed by the Voronoi-based method. This result highlights the efficiency of our framework in optimizing infrastructure deployment. Our reinforcement learning-based method for charging station placement is particularly noteworthy. At each step of the placement process, it achieves optimal performance by effectively maximizing future rewards in accordance with reinforcement learning theory. In other words, every decision made not only addresses current demand but also strategically positions the infrastructure for long-term efficiency and benefits.

\subsubsection{Charging Station Sizing Index (CSSI)}
\begin{figure}[!t]
    \centering
    \includegraphics[width= 0.8\linewidth]{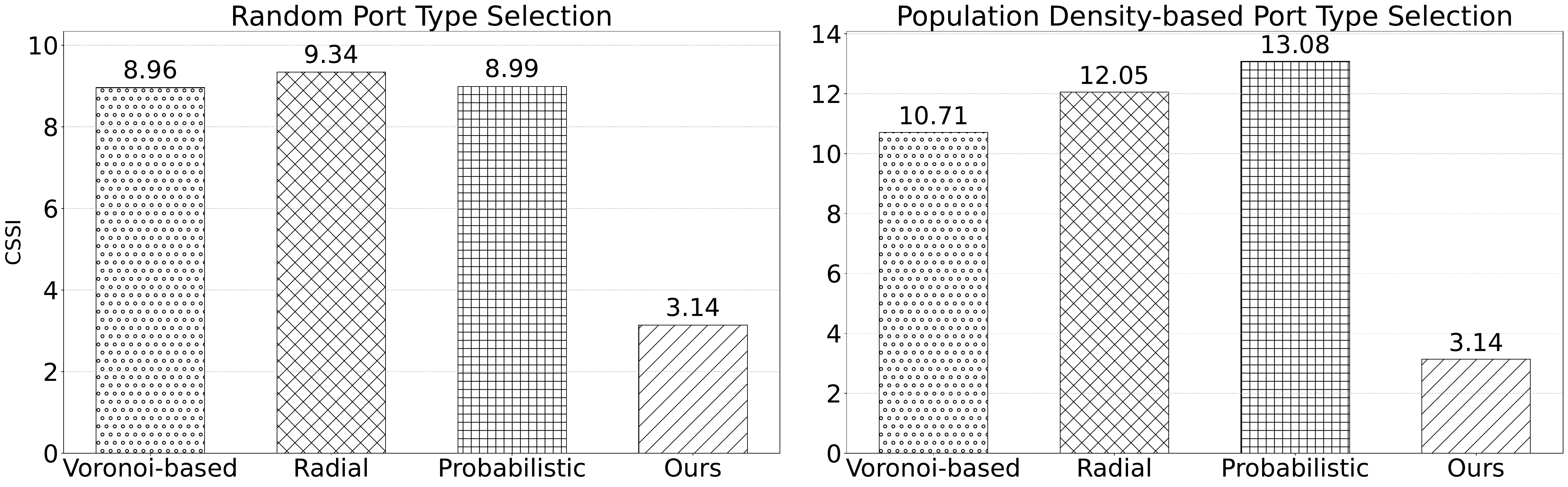}
    \caption{Comparison of charging station sizing index (CSSI) across different station placement and port selection baseline methods and ours.}
    \label{fig: sizing}
\end{figure}
To quantitatively assess the scale of a network of charging stations, we introduce the \emph{Charging Station Sizing Index} (CSSI).  Let there be \(N\) stations and ten port types (indexed by \(j\)) and an associated scaling factor \(s_j\) that reflects its relative capacity. We assign $s_i$ an increasing value from $0.1$ to $1$ in the order of the port types in Table \ref{tab: quantity}. A larger $s_i$ corresponds to a larger station scale:

At station \(i\), let \(p_{i,j}\) be the number of ports of type \(j\).  We define
\begin{equation}
\mathrm{CSSI}
= \frac{1}{N}\sum_{i=1}^{N}\sum_{j=1}^{10} pt_{i,j}\,s_j,
\end{equation}
where $N$ is the total number of stations, $pt_{i,j}$ is the count of type-$j$ ports at station $i$, $s_j$ is the scaling factor for port type $j$.

By construction, \(\mathrm{CSSI}\) serves as a scaling coefficient for the charging station system. A higher \(\mathrm{CSSI}\) indicates a larger scale of the system, while a lower \(\mathrm{CSSI}\) suggests a smaller scale for the charging station network capability.

We present the results in Figure~\ref{fig: sizing}, which show that our framework achieves a relatively low score. This outcome indicates that the proposed infrastructure is effectively aligned with actual demand while optimizing system performance. In the left panel, we observe that baselines with random port type selection exhibit an equal distribution. In contrast, using population density-based port type selection results in a higher value for the probabilistic method. However, these baselines require a large-scale system to achieve optimal performance.

\section{Conclusion}
Our study introduces a pioneering framework for optimizing electric vehicle charging station placement by integrating deep reinforcement learning with GAMA-based simulations. Addressing the shortcomings of traditional deterministic approaches, our method dynamically models EV movement and obtains the charging demand, enabling adaptive and efficient infrastructure planning. By employing a hybrid RL agent with dual deep Q-Networks for location selection and port configuration, we effectively balance critical factors such as population density, proximity to existing stations, and substation availability, guided by simulation-derived rewards.

Experimental results, conducted in a simulated environment based on Hanoi, Vietnam, demonstrate the framework’s superiority, achieving a 53.28\% reduction in average waiting times compared to the original state and outperforming static baseline methods. This significant improvement underscores the value of incorporating real-time feedback and flexible reward structures over static optimization techniques. Additionally, the framework prioritizes station placement in high-demand areas near existing infrastructure, ensuring efficient resource distribution and user convenience, as evidenced by the reduced mean distance to existing stations.

Furthermore, our analysis of future EV demand highlights the scalability of the proposed method, demonstrating optimal performance with fewer charging stations compared to baseline approaches. The introduction of the Charging Station Sizing Index (CSSI) provides a quantitative measure of infrastructure scale, revealing that our framework achieves a lower CSSI, indicating alignment with actual demand while maintaining high system efficiency. These results underscore the superiority of integrating simulation-based rewards and RL over static optimization techniques, particularly in predicting the necessary scale and placement of charging stations to meet both current and future needs.

While our framework shows promising results, we acknowledge that any simulation is a simplification of reality. To enhance generalizability, the behavioral parameters in our GAMA model were defined based on expert surveys and validated against real-world charging data. This ensures realistic patterns in EV movement, queuing, and charging behaviors. Furthermore, the modularity of the framework allows user behavior models to be refined or replaced with real-world EV usage data as it becomes available. We argue that learning from simulated demand dynamics provides a more transferable signal than purely deterministic rewards, which are prone to overfitting to static historical datasets.

Regarding scalability, our framework is designed to extend to larger urban areas. As the number of existing stations and candidate locations increases, the RL agent encounters a more diverse set of states and scenarios during training, naturally enhancing the generality of the learned policy. This means scaling the problem size strengthens rather than limits the framework's effectiveness. For generalizability to new contexts, the data-driven RL agent can be retrained on local geospatial data (e.g., population density, traffic patterns, power grid infrastructure), while the GAMA simulation can be recalibrated with region-specific parameters to reflect distinct EV user behaviors, ensuring extensibility and transferability.

Our reward function inherently balances key trade-offs, such as minimizing average waiting time (through the simulation-derived $r_{wait}$ component, which promotes efficiency by reducing congestion) and maintaining equitable access to EV charging infrastructure across geographic regions via the distance-to-existing-stations reward $r_{exist}$ and the Voronoi-based action space, which prioritize underserved areas and promote spatial coverage.
Despite these strengths, limitations remain. For instance, we adopted equal weights for reward components as a principled baseline to avoid subjective bias, given the computationally prohibitive nature of exhaustive searches like grid search in our integrated RL-simulation setup; each weight configuration would require full retraining and re-simulation cycles. 

In future work, we plan to address these by exploring advanced techniques deep learning for efficient reward weight tuning. Additionally, we intend to recalibrate simulation parameters for new environments, incorporate probabilistic load forecasting to complement our siting focus, and conduct further ablation studies to quantify component impacts. These extensions will further refine our framework, contributing to sustainable urban mobility and intelligent transportation systems.


\section*{Acknowledgement}
This work was supported by the Center for Environmental Intelligence, VinUniversity [grant number CEI Flagship  VUNI.CEI.FS\_0006 and  VUNI.CEI.FS\_0007]



\bibliographystyle{elsarticle-harv}
 \bibliography{cas-refs}





\end{document}